\documentclass[journal]{IEEEtran}
\usepackage{epstopdf}
\usepackage{epsfig}
\usepackage{times}
\usepackage{epsfig}
\usepackage{graphicx}
\usepackage{subfigure}
\usepackage{amsmath}
\usepackage{multirow}
\usepackage{amssymb}
\usepackage{algorithm}
\usepackage{algorithmic}
\usepackage{multirow}
\usepackage{color}
\ifCLASSINFOpdf
\else

\fi
\begin{document}
\title{Part-based Deep Hashing for Large-scale \\Person Re-identification$^{*}$}

\author{Fuqing~Zhu,
        Xiangwei~Kong,~\IEEEmembership{Member,~IEEE,}
        Liang~Zheng,~\IEEEmembership{Member,~IEEE,}
        Haiyan~Fu,~\IEEEmembership{Member,~IEEE,}
        Qi~Tian,~\IEEEmembership{Fellow,~IEEE,}
\thanks{This work was supported in part by the Foundation for Innovative Research Groups of the National Natural Science Foundation of China (NSFC) under Grant 71421001, in part by the National Natural Science Foundation of China (NSFC) under Grant 61502073 and Grant 61429201, in part by the Open Projects Program of National Laboratory of Pattern Recognition under Grant 201407349, and in part to Dr. Qi Tian by ARO grants W911NF-15-1-0290 and the Faculty Research Gift Awards by NEC Laboratories of America and Blippar. \emph{(Corresponding author: Xiangwei Kong)}}
\thanks{F. Zhu, X. Kong and H. Fu are with the School of Information and Communication Engineering, Dalian University of Technology, Dalian 116024, China (e-mail: fuqingzhu@mail.dlut.edu.cn; kongxw@dlut.edu.cn; fuhy@dlut.edu.cn).}
\thanks{L. Zheng is with the University of Technology Sydney, NSW, Australia. (e-mail: liangzheng06@gmail.com).}
\thanks{Q. Tian is with the University of Texas at San Antonio, Texas 78249-1604,
USA (e-mail: qitian@cs.utsa.edu).}
\thanks{$^{*}$Citation for this paper: F. Zhu, X. Kong, L. Zheng, H. Fu, and Q. Tian, ``Part-based Deep Hashing for Large-scale Person Re-identification,'' \emph{IEEE Transactions on Image Processing}, DOI: 10.1109/TIP.2017.2695101, 2017.}}
\markboth{IEEE TRANSACTIONS ON IMAGE PROCESSING}%
{Shell \MakeLowercase{\textit{et al.}}: Bare Demo of IEEEtran.cls for IEEE Journals}

\maketitle

\begin{abstract}
Large-scale is a trend in person re-identification (re-id). It is important that real-time search be performed in a large gallery. While previous methods mostly focus on discriminative learning, this paper makes the attempt in integrating deep learning and hashing into one framework to evaluate the efficiency and accuracy for large-scale person re-id. We integrate spatial information for discriminative visual representation by partitioning the pedestrian image into horizontal parts. Specifically, Part-based Deep Hashing (PDH) is proposed, in which batches of triplet samples are employed as the input of the deep hashing architecture. Each triplet sample contains two pedestrian images (or parts) with the same identity and one pedestrian image (or part) of the different identity. A triplet loss function is employed with a constraint that the Hamming distance of pedestrian images (or parts) with the same identity is smaller than ones with the different identity. In the experiment, we show that the proposed Part-based Deep Hashing method yields very competitive re-id accuracy on the large-scale Market-1501 and Market-1501+500K datasets.
\end{abstract}

\begin{IEEEkeywords}
Deep Learning, hashing, part-based, large-scale person re-identification.
\end{IEEEkeywords}

\IEEEpeerreviewmaketitle

\section{Introduction}
\IEEEPARstart{T}{his} paper focuses on large-scale person re-identification (re-id), which has received increasing attention in automated surveillance for its potential applications in human retrieval, cross-camera tracking and anomaly detection. Given a pedestrian image, person re-id aims to match in a cross-camera database for the bounding boxes that contain the same person. Matching cross scenarios is challenging due to the varieties of lighting, pose and view point.

Person re-id lies in between image classification \cite{yan2016image,yang2013feature,chang2016compound} and retrieval \cite{yang2008harmonizing,yang2012multimedia}, which has made detailed discussion in \cite{Zheng2016Personi}. Previous person re-id works \cite{li2014deepreid,liao2015person,zheng2016mars,zheng2015scalable} usually take advantage of both image classification and retrieval. This work considers two issues in large-scale person re-id: efficiency and CNN models for effective descriptors. On the one hand, computational efficiency has been a concern in person re-id works. Some state-of-the-art methods employ brute-force feature matching strategies \cite{zhao2013person,zhao2013unsupervised}, which obtain good matching rate. However, these methods suffer from low computational efficiency in large-scale applications. Motivated by \cite{zheng2015query,zheng2015scalable}, we view person re-id as a special task of image retrieval. Both tasks share the same target:  finding the images containing the same object/pedestrian as the query \cite{zheng2015scalable}. A reasonable choice to address the above efficiency problem of large-scale person re-id therefore involves the usage of image retrieval techniques. Hashing, known for fast Approximate Nearest Neighbor (ANN) search, is a good candidate in our solution kit. The main idea of hashing method is to construct a series of hash functions to map the visual feature of image into a binary feature vector so that visually similar images are mapped into similar binary codes. Recently, hashing methods based on deep neural networks \cite{xia2014supervised,zhao2015deep,lai2015simultaneous,lin2015deep,zhang2015bit,lai2016instance} obtain higher accuracy than traditional hashing methods. However, to our knowledge, there are few works employing hashing to address large-scale person re-id.

On the other hand, the Convolutional Neural Network (CNN) has demonstrated its effectiveness in improving accuracy of person re-id \cite{ahmed2015improved,chen2015deep,li2014deepreid,zheng2016mars}. The Siamese CNN model uses training image pair as input and a binary classification loss is used to determine if they belong to the same ID. This cross-image representation is effective in capturing the relationship between the two images and addressing horizontal displacement problem. For the conventional classification based CNN model, Zheng \emph{et al.} \cite{zheng2016mars} propose to learn an ID-discriminative embedding to discriminate between pedestrians in the testing set. These methods, while achieving impressive person re-id accuracy, do not address the efficiency issue either, because they typically use the Euclidean or Cosine distance for similarity calculation which is time-consuming under large galleries and high feature dimensions. Currently the largest person re-id dataset Market-1501 \cite{zheng2015scalable} contains 32,668 annotated bounding boxes, plus a distractor set of 500K images. It poses the scaling problem for person re-id methods. This paper therefore investigates how to balance re-id effectiveness and efficiency.

The approach we pursue in this work, as mentioned above, is motivated by hashing and CNN, which takes into account the efficiency and accuracy, respectively. A  triplet loss based supervised Deep Hashing framework is employed to address the efficiency of large-scale person re-id. The triplet deep neural networks \cite{hoffer2015deep}, \cite{schroff2015facenet}, \cite{wang2014learning}, which have been used in face recognition \cite{schroff2015facenet} and fine-grained image similarity models \cite{wang2014learning}, learn discriminative embeddings by imposing a relative distance constraint. The relative distance constraint aims to minimize the distance between positive pairs, while pushing away the negative pairs. This constraint is flexible comparing with restricting the distances of positive or negative pairs in an absolute range. Moreover, the spatial information of pedestrian image is beneficial for higher person re-id accuracy, because the local parts of pedestrians provide more precise matching strategy compared with using the entire pedestrian images. The part-based trick is useful for improve the accuracy in face verification, such as DeepID \cite{sun2014deep} and DeepID2 \cite{sun2014deepid2}. In DeepID \cite{sun2014deep}, the face image is converted into ten parts which are global regions taken from the weakly aligned faces and local regions centered around the five facial landmarks, respectively. However, the part partitioning strategies of DeepID is not suitable for ensuring the efficiency of large-scale person re-id. For simplicity, in this paper we just partition the entire pedestrian image into horizontal 3 or 4 parts without any semantic alignment strategy. Our work gives two aspects of improvement on the basis of triplet-based deep neural network works \cite{hoffer2015deep}, \cite{wang2014learning} for large-scale person re-id. First, in the intermediate layers of CNN, a hash layer is designed to make the output of network suitable for binarization. Second, the proposed network is composed by several sub-network branches for individual parts, and each sub-network branch is a triplet-based deep network. From the above consideration, we propose a \textbf{P}art-based \textbf{D}eep \textbf{H}ashing (PDH) method for large-scale person re-id. Our goal is to generate a binary representation for each pedestrian image using the deep CNN, which 1) is effective in discriminate different identities, 2) integrates spatial constraint, and 3) improves efficiency for the large-scale pedestrian gallery in terms of both memory and speed. Our code will be available at the website https://sites.google.com/site/fqzhu001.

Different from most previous works on person re-id, this paper focuses on hashing methods on Market-1501 dataset and its associating distractor set with 500K images. To our best knowledge, there is only one published paper which utilizes deep hashing for person re-id \cite{zhang2015bit} on CUHK 03 \cite{li2014deepreid}, a dataset having  only  100 identities in each gallery split. We show that our method yields effective yet efficient person re-id performance compared to  several competing methods. The main contributions of this paper are listed below.

\begin{itemize}
\item Among the first attempts, we employ hashing to improve the efficiency for large-scale person re-id. While several previous works \cite{zhang2015bit} only use small datasets, this paper reports large-scale evaluation results on the largest Market-1501 and Market-1501+500K datasets, such gaining more insights into the hashing task. The binary hash codes achieve fast matching for large-scale person re-id, which addresses the problem of computational and storage efficiency.
\item A part-based model is integrated into the deep hashing framework to increase the discriminative ability of visual matching. The performance increases significantly compared with the baseline.
\end{itemize}

The rest of the paper is organized as follows. In section \ref{sect:2}, we review related work briefly. The proposed PDH method will be described in section \ref{sect:3}. In section \ref{sect:4}, extensive results are presented on Market-1501 and Market-1501+500K datasets. Finally, we conclude the paper in section \ref{sect:5}.

\textbf{Note:} this work was done in late 2015 when we were trying the triplet loss network to learn embeddings. Later, we turn to the identification models and obtain more competitive results. Interested readers can also refer to our works on the identification models \cite{zheng2016person,zheng2016mars,Zheng2016Personi,zheng2017unlabeled,lin2017improving,sun2017svdnet}.

\section{Related Work}\label{sect:2}
This paper considers the efficiency and accuracy of large-scale person re-id via deep hashing method. So we briefly review the methods of person re-id using both hand-crafted and deeply-learned features, and hashing methods.

\subsection{Hand-crafted Methods for Person Re-identification}
The previous mainstream works in  person re-id typically focus on visual feature representation \cite{zhao2013person,zheng2015query,su2015multi} and distance metric learning \cite{zhao2014learning,shen2015person,prosser2010person}. On feature representation, Ma \emph{et al}. \cite{ma2012bicov} utilize Gabor filters and Covariance descriptors to deal with illumination changes and background variations, while Bazzani \emph{et al}. \cite{bazzani2014sdalf} design a Symmetry-Driven Accumulation of Local Features (SDALF) descriptor. Inspired by recent advanced Bag-of-Words (BOW) model in large-scale image retrieval field, Zheng \emph{et al}. \cite{zheng2015scalable} propose an unsupervised BOW based descriptor. By generating a codebook on training data, each pedestrian image is represented as a histogram based on visual words. Li \emph{et al}. \cite{li2015cross} learn a cross-view dictionaries based on SIFT and color histogram to obtain an effective patch-level feature across different views for person re-id. Ma \emph{et al}. \cite{ma2012local} use Fisher Vector (FV) to encode local feature descriptors for patches to improve the performance of person re-id. Liao \emph{et al}. \cite{liao2015person} propose a method for building a descriptor which was invariant to illumination and viewpoint changes. Zhao \emph{et al}. \cite{zhao2013person} propose a method which assigned different weights to rare colors on the basis of salience information among pedestrian images. However, traditional fixed hand-crafted visual features may not optimally represent the visual content of images. That means a pair of semantically similar pedestrian images may not have feature vectors with relatively small Euclidean distance. In the work of distance metric learning methods for person re-id, the classic RankSVM \cite{prosser2010person,zhao2014learning} and boosting \cite{shen2015person} methods are widely used. B. Prosser \emph{et al}. \cite{prosser2010person} solve person re-id task as a ranking problem using RankSVM to learn similarity parameters. The method of KISSME \cite{koestinger2012large} and EIML \cite{hirzer2012person} are effective metric learning methods which have been shown in \cite{roth2014mahalanobis}.

\subsection{Deeply-learned Methods for Person Re-identification}
Recently the state-of-the-art methods in person re-id have been dominated with deep learning models. The main advantage is that the CNN framework can either optimize the feature representation alone \cite{zheng2016mars} or simultaneously learn features and distance metrics \cite{li2014deepreid}. Li \emph{et al}. \cite{li2014deepreid} propose a filter pairing neural network (FPNN) by a patch matching layer and a maxout-grouping layer. The patch matching layer is used to learn the displacement of horizontal stripes in across-view images, while the maxout-grouping layer is used to boost the robustness of patch matching. Ahmed \emph{et al}. \cite{ahmed2015improved} design an improved deep neural network by adding a special layer to learn the cross-image representation via computing the neighborhood distance between two input images. The softmax classifier is added on the learned cross-image representation for person re-id. Yi \emph{et al}. \cite{yi2014deep} employ the Siamese architecture which consists of two sub-networks. Each sub-network processes one image independently and the final representations of images are connected to evaluate similarity by a special layer. The deep networks are trained by preserving the similarity of the two images. The author evaluates the performance on VIPER \cite{gray2008viewpoint} and PRID-2011 \cite{hirzer2011person} datasets. However, the VIPER and PRID-2011 are both comparatively small datasets.  E. Ustinova \emph{et al}. \cite{ustinova2015multiregion} utilize bilinear pooling method based on Bilinear CNN for person re-id, which is implemented over multi-region for extracting more useful descriptors in the two large datasets CUHK 03 \cite{li2014deepreid} and Market-1501 \cite{zheng2015scalable}. Chen \emph{et al}. \cite{chen2015deep} design a deep ranking framework to formulate the person re-id task. The image pair is converted into a holistic image horizontally firstly, then feeds these images into CNN to learn the representations. Finally the ranking loss is used to ensure that positive matched image pair is more similar than negative matched image pair. Wang \emph{et al}. \cite{wang2016joint} design a joint learning deep CNN framework, in which the matching of single-image representation and the classification of cross-image representation are jointly optimized for pursuing better matching accuracy with moderate computational cost. Since single-image representation is efficient in matching, while cross-image representation is effective in modeling the relationship between probe image and gallery image, the fusion of two representation losses together is utilized the advantages of both these representations. Xiao \emph{et al}. \cite{Xiao2016Learning} propose a pipeline for learning generic and robust deep feature representations from multiple domains with CNN, in which the Domain Guided Dropout algorithm is utilized to improve the feature learning procedure.

\subsection{Review of Hashing Methods}
The field of fast Approximate Nearest Neighbor (ANN) search has been greatly advanced due to the development of hashing technique, especially those based on deep CNN. For the non-deep hashing methods, the hash code generation process has two stages. First, the image is represented by a vector of hand-crafted visual features (such as Gist descriptor). Then, separate projection or quantization step is used to generate hash codes. Unsupervised and supervised hashing are two main streams, such as Spectral Hashing (SH) \cite{Weiss2008Spectral}, Iterative Quantization (ITQ) \cite{Gong2011Iterative}, Semi-supervised Hashing (SSH) \cite{Wang2010Semi}, Minimal Loss Hashing (MLH) \cite{Norouzi2011Minimal}, Robust Discrete Spectral Hashing (RDSH) \cite{yang2015robust}, Zero-shot Hashing (ZSH) \cite{Yang2016Zero} and Kernel Supervised Hashing (KSH) \cite{Liu2012Supervised}. However, hashing methods based on hand-crafted features may not be effective in dealing with the complex semantic structure of images, thus producing sub-optimal hash codes.

The deep hashing method maps the input raw images to hash codes directly, which learns feature representation and the mapping from the feature to hash codes jointly. Xia \emph{et al}. \cite{xia2014supervised} propose a supervised deep hashing method CNNH, in which the learning process is decomposed into a stage of learning approximate hash codes from similarity matrix, followed by a stage of simultaneously learning hashing functions and image representations based on the learned approximate hash codes. Zhao \emph{et al}. \cite{zhao2015deep} propose a Deep Semantic Ranking Hashing (DSRH) method to employ multi-level semantic ranking supervision information to learn hashing function, which preserves the semantic similarity between multi-label images. Lai \emph{et al}. \cite{lai2015simultaneous} develop a ``one-stage'' supervised hashing framework by a well designed deep architecture. The deep neural network employs the shared sub-network which makes feature learning and hash coding process simultaneously. Lin \emph{et al}. \cite{lin2015deep} propose a point-wise supervised deep hashing method by adding a latent layer in the CNN for fast image retrieval. Zhang \emph{et al}. \cite{zhang2015bit} propose a novel supervised bit-scalable deep hashing method for image retrieval and person re-id. By designing an element-wise layer, the hash codes can be obtained bit-scalability, which is more flexible to special task when need different length of hash codes.

\begin{figure}[t]
\centering
\includegraphics[width=0.485\textwidth{},keepaspectratio]{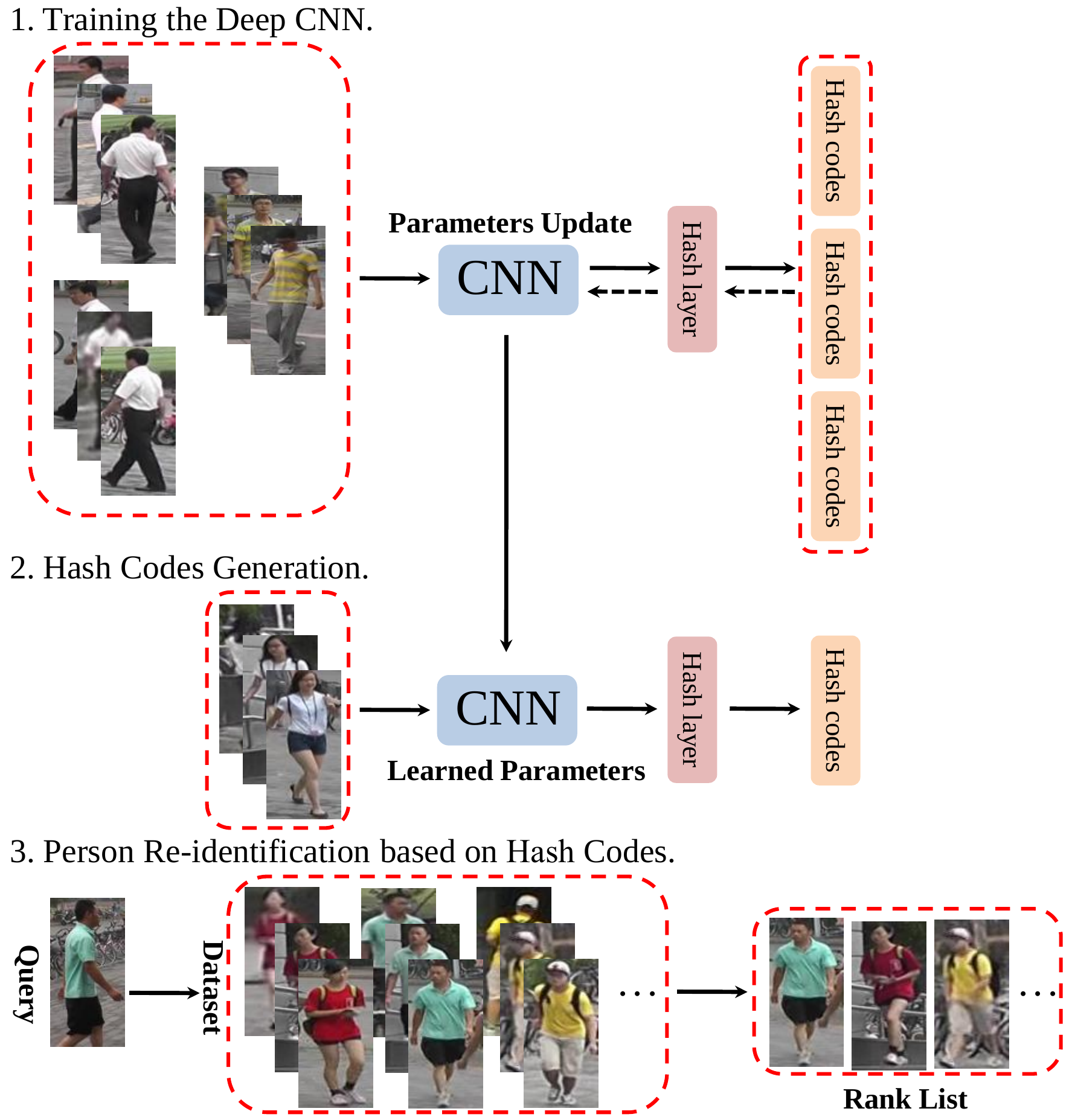}%
\caption{The overview of baseline triplet-based deep hashing model for large-scale Person Re-id. The deep hashing model consists of three main modules. The first module is training the deep CNN model. In the second module, the trained CNN model is used to generate hash codes for pedestrian images of query and testing set. The final module is to retrieve similar images based on Hamming distance of hash codes between the samples of query and testing set.}\label{fig:base_g}
\end{figure}

It is true that deep hashing has been employed in image retrieval, and that part-based method is a common technique to improve re-id performance. However, both techniques have rarely been evaluated in person re-id and hashing tasks, respectively, especially in large-scale settings. Our work departs from previous person re-id works. We apply such simple yet effective techniques on the Market-1501 and Market-1501+500K datasets, and provide insights on how re-id performance (efficiency and accuracy) can be improved on the large-scale settings.

\section{Proposed Approach}\label{sect:3}
\begin{figure*}[t]
\centering
\includegraphics[width=0.9\textwidth{},keepaspectratio]{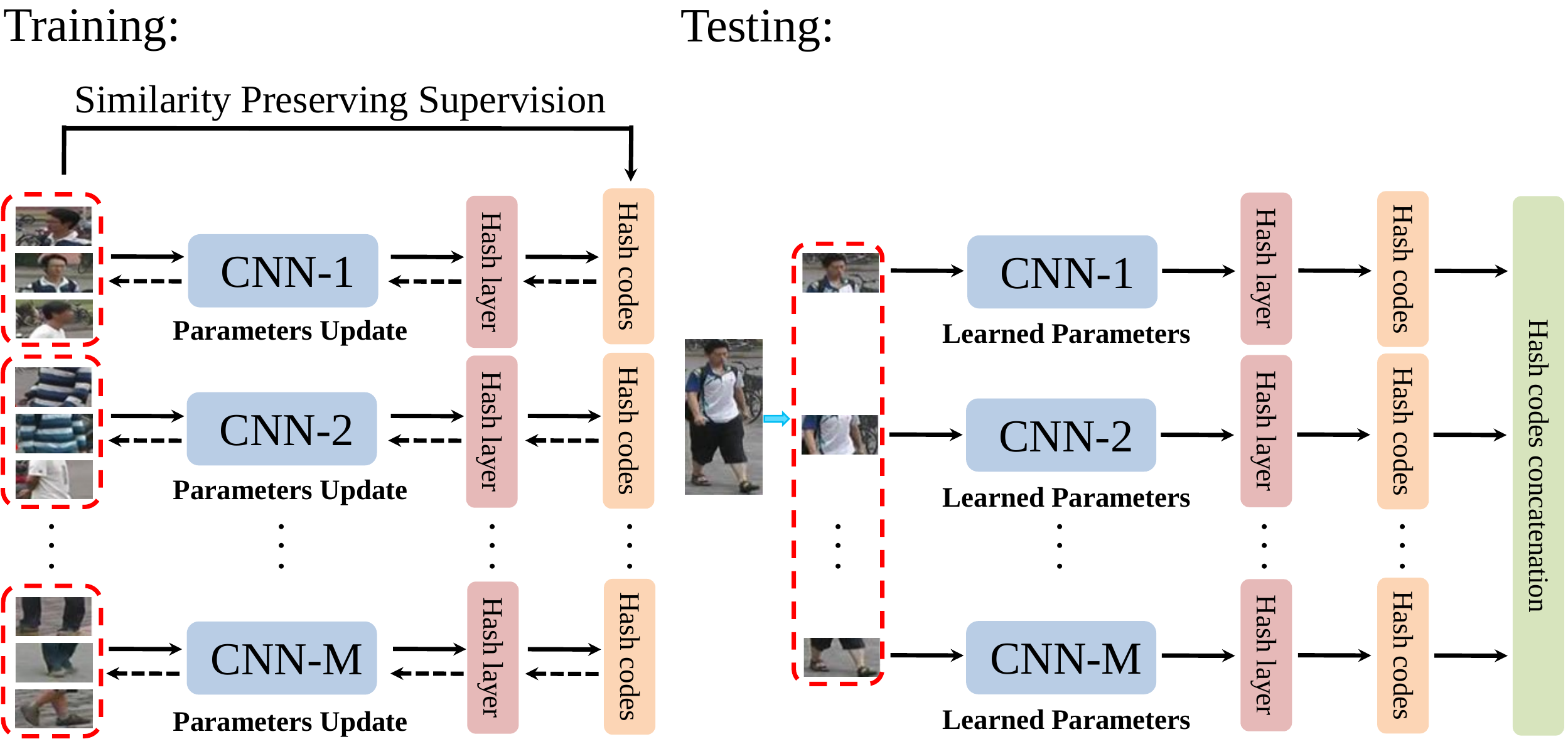}
\caption{The framework of the proposed Part-based Deep Hashing (PDH) model.}\label{fig:part_m}
\end{figure*}

The task of person re-id is to match relevant pedestrian images for a query in the cross-camera scenario. Due to the variation of pedestrian in different scenarios, the spatial information is important for enhancing the discriminative ability of image representation. This is the motivation of integrating part-based model into the baseline triplet-based deep hashing framework, so that more discriminative hash codes can be generated. First, an overview of the baseline triplet-based deep Convolutional Neural Network (CNN) hashing framework for person re-id is illustrated in Fig. \ref{fig:base_g}. The triplet-based deep CNN hashing framework to generate the binary hash codes for pedestrian images based on the CaffeNet \cite{krizhevsky2012imagenet}, where a hash layer is well designed to ensure the compact binary output. In the training phase, a triplet-based loss function is employed for learning optimal parameters of the deep CNN model. Second, the proposed PDH method is implemented on the basis of the triplet-based deep hashing framework, which is illustrated in Fig. \ref{fig:part_m}. For the part subsets of pedestrian images at the same corresponding location, we train a separate network for each part subset  and obtain a series of optimal part-based deep CNN models. In this way, the corresponding parts of testing pedestrian image are processed by a series of trained part-based deep CNN models. The final representation of pedestrian image is the concatenation of each part result. The learned hash codes will be directly used for person re-ID without any feature selection \cite{Chang2016Convex,chang2016semisupervised} process. Third, due to each identity has multiple query images in a single camera, multiple query images are merged into a single query for a further accuracy improvement of large-scale person re-id.

\subsection{Baseline Triplet-based Deep Hashing Model}\label{sec:base_l}
We employ the triplet-based deep hashing method to solve the efficiency problem of large-scale person re-id. The baseline method of triplet-based deep hashing is an end-to-end framework which jointly optimizes the image feature representation and hashing function learning, \emph{i.e.} the input of framework is raw pixels of pedestrian images, while the output is hash codes. For the task of person re-id, the aim is to obtain the hash code of pedestrian image by the trained deep hashing model. How to train a discriminative deep neural network that can preserve the similarity of samples is critical. We briefly describe the training process of triplet-based deep CNN hashing model.
\begin{figure}[t]
\centering
\includegraphics[width=0.45\textwidth{},keepaspectratio]{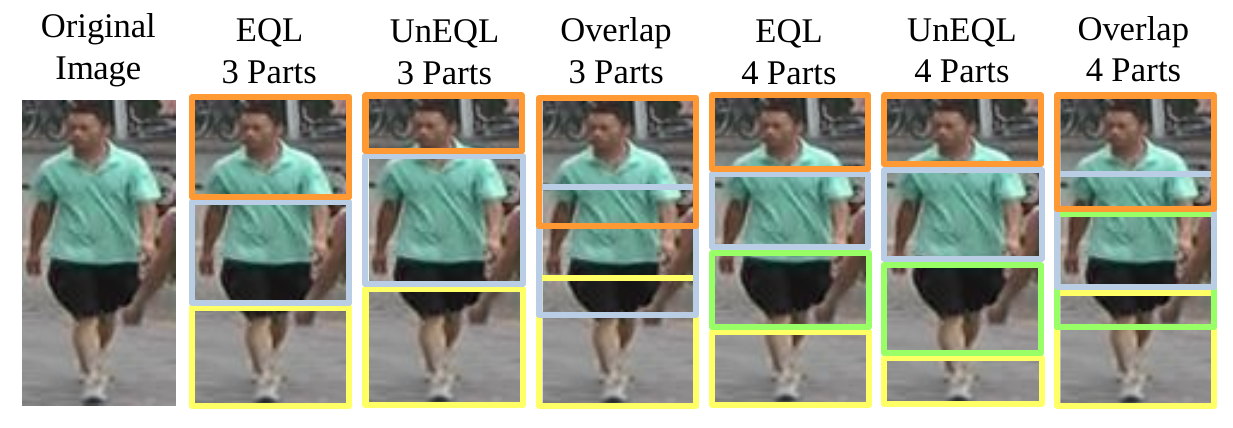}%
\caption{Different part partitioning strategies. ``EQL'' and ``UnEQL'' represent dividing the image equally and unequally, respectively.}\label{par_s}
\end{figure}

Each training sample is associated with an identity label. The principle of learning optimal deep neural network is formulated to ensure the Hamming distance of hash codes small for same identity samples. Meanwhile, the Hamming distance of binary hash codes should be large for different identity samples. The triplet-based input form is suitable for learning the parameters of deep neural network. Each triplet input includes three pedestrian images, in which one of them is anchor. The other two images are the same and different identity samples with the anchor, respectively.

Let $I_i$ be anchor. $I_i^{+}$ and $I_i^{-}$ are the same and different identity samples with the anchor, respectively. Let the hash code representation of image $I_i$ represent as $H(I_i)$, which is the response of hash layer. The hash layer follows fully connected layer ($FC7$).
\begin{equation}H\left( {{I_i}} \right) = {sign}\left( {{{\bf{w}}^T}FC7\left( {{I_i}} \right)}\right),
\end{equation}
where ${\bf{w}}$ denotes weights in the hash layer and $ sign \left(  v  \right)$ returns $1$ if $v > 0$ and $0$ otherwise. According to this criterion, the objective function is
\begin{equation}\mathop {\min }_{{\bf{W}}} \sum\limits_{i = 1}^N {\left( {{{\left\| {H({I_i}) - H(I_i^ {+} )} \right\|}_H} - {{\left\| {H({I_i}) - H(I_i^ {-} )} \right\|}_H}} \right)}, \end{equation}
where $\bf{W}$ denotes weights of each layer. The weights updating of each layer is achieved by a triplet-based loss function which is defined by \begin{equation}\begin{split}& \mathcal{L}\left( H(I_i),H({I_i^{+}}),H({I_i^{-}}) \right)\\&=\max ( 0,1-({{\| H(I_i)-H({I_i^{-}}) \|}_{H}}\\&-{{\| H(I_i)-H({I_i^{+}}) \|}_{H}}) ),\end{split}\end{equation}
where $H(I_i),H({I_i^{+}}),H({I_i^{-}})\in {{\left\{ 0,1 \right\}}^{q}}$ and ${{\left\| \cdot  \right\|}_{H}}$ represents the Hamming distance. The loss function (3) is not differentiable due to the ${{\left\| \cdot  \right\|}_{H}}$ of (3) and the $sign$ function of (1). To facilitate the optimization, a relaxation trick on $(3)$ is utilized to replace the Hamming distance with the $L_{2}$ norm. In addition, we replace the $sign$ function of (1) with $sigmoid$ function. Let ${f_H}\left( {{I_i}} \right)$ represent the relaxation of $H(I_i)$.
\begin{equation}{f_H}\left( {{I_i}} \right) = {sigmoid}\left( {{{\bf{w}}^T}FC7\left( {{I_i}} \right)}\right), \end{equation}
where $sigmoid$ function is defined as: \begin{equation}sigmoid\left( x \right) = \frac{1}{{1 + {e^{ - x}}}}.\end{equation} The $sigmoid$ function can restrict the output value ${f_H}\left( {{I_i}} \right)$ in the range $[0,1]$. The modified loss function is\begin{equation}\begin{split}& \mathcal{L}\left( f_H(I_i),f_H({I_i^{+}}),f_H({I_i^{-}}) \right)\\&=\max ( 0,1-({{\| f_H(I_i)-f_H({I_i^{-}}) \|}_{2}}\\&-{{\| f_H(I_i)-f_H({I_i^{+}}) \|}_{2}}) )\end{split}\end{equation}
where $f_H(I_i),f_H({I_i^{+}}),f_H({I_i^{-}})\in {{\left[ 0,1 \right]}^{q}}$.

In this way, the variant of triplet loss becomes a convex optimization problem. If the condition
$1-({{\left\| f_H(I_i)-f_H({I_i^{-}}) \right\|}_{2}}-{{\left\| f_H(I_i)-f_H({I_i^{+}}) \right\|}_{2}})>0$ is satisfied, their gradient values are as follows:
\begin{equation}\begin{split}&\frac{\partial \mathcal{L}}{\partial f_H(I_i)}=2\left( f_H({I_i^{-}})-f_H({I_i^{+}}) \right)\\&\frac{\partial \mathcal{L}}{\partial f_H({I_i^{+}})}=2\left( f_H({I_i^{+}})-f_H(I_i) \right)\\&\frac{\partial \mathcal{L}}{\partial f_H({I_i^{-}})}=2\left( f_H({I_i^{-}})-f_H(I_i) \right).\end{split}\end{equation}
These gradient values can be fed into the deep CNN by the back propagation algorithm to update the parameters of each layer.

After the deep neural network model is trained, the new input pedestrian image $I_j$ in query and testing set can be evaluated to generate hash code in the testing phase. The final binary representation of each image $I_j$ is $\hat{H}(I_j)$, which is operated by simple quantization:
\begin{equation}\hat{H}\left( {{I_j}} \right) = {sign}\left({f_H}\left( {{I_j}} \right)- 0.5 \right).
\end{equation}

\setlength{\tabcolsep}{20.5pt}
\begin{table}[t]
\centering
\caption{The Size of Parts of Various Region Partitioning Methods. ``EQL'' and ``UnEQL'' Represent Dividing the Image Equally and Unequally, Respectively. }
\begin{tabular}{l|c}
\hline Region Partitions&Size\\
\hline EQL 3 Parts&42$\times$64; 42$\times$64; 42$\times$64;\\
 UnEQL 3 Parts&24$\times$64; 56$\times$64; 48$\times$64;\\
 Overlap 3 Parts&56$\times$64; 56$\times$64; 56$\times$64;\\
 EQL 4 Parts&32$\times$64; 32$\times$64; 32$\times$64; 32$\times$64;\\
 UnEQL 4 Parts&28$\times$64; 40$\times$64; 40$\times$64; 20$\times$64;\\
 Overlap 4 Parts&48$\times$64; 48$\times$64; 48$\times$64; 48$\times$64;\\
\hline
\end{tabular}\label{table:part_size}
\end{table}

\subsection{The Proposed Part-based Deep Hashing Model}\label{sec:part_model}
Due to the intensely variation of pedestrian in cross-camera scenarios, the spatial information of the pedestrian image is significant for enhancing the discriminative ability. A logical idea is to utilize the local part instead of the entire image to train the deep model. According to the consistency of person spatial information, we briefly make 6 part partitioning variants, which are listed in Table \ref{table:part_size}. The direction of region partition is along with horizontal and from top to bottom. The examples of different part partitioning are shown in Fig. \ref{par_s}. The size of parts of various region partitioning methods is shown in Table \ref{table:part_size}. We can train deep hashing model for each part separately instead of entire image. However, we do not know which part of the pedestrian image is more beneficial for training the deep hashing model. A simple strategy is to combine the results of each part with a uniform standard. To avoid complex calculation, we just divide the pedestrian image into a few parts. The number of the part for a pedestrian image and the trained deep CNN models is consistent. The architecture of proposed PDH method is shown in Fig. \ref{fig:part_m}, which is on the basis of baseline triplet deep hashing model. The PDH method is as follows:

In the Training Phase, first, the training pedestrian image $I_i$ is divided into a few parts. \emph{i.e.}
${I_i} = \left\{ {{I_{i,k}}} \right\},k = 1,...,M$, where $I_{i,k}$ is the $k$-th part of pedestrian image $I_i$ and $M$ is the number of parts of one image.

Then, the same locations of pedestrian images constitute a specific part-based subset. The number of training samples is $N$. The total number of subset is $M$. The $k$-th subset is denoted as:

\begin{equation}
{\rm subset} = \{ {I_{i,k}}\} ,i = 1,...,N.
\end{equation}

Finally, for each subset, we train the deep CNN model using the samples of subset, and obtain the learned parameters of each layers. The loss function is as follows:
\begin{equation}\begin{split}& \mathcal{L}\left( f_H(I_{i,k}),f_H({I_{i,k}^{+}}),f_H({I_{i,k}^{-}}) \right)\\&=\max ( 0,1-({{\| f_H(I_{i,k})-f_H({I_{i,k}^{-}}) \|}_{2}}\\&-{{\| f_H(I_{i,k})-f_H({I_{i,k}^{+}}) \|}_{2}}))\end{split}\end{equation}
where $f_H(I_{i,k}),f_H({I_{i,k}^{+}}),f_H({I_{i,k}^{-}})\in {{\left[ 0,1 \right]}^{q}},k=1,...,M.$, and ${f_H}\left( {{I_{i,k}}} \right)={sigmoid}\left({\bf{w}}_k^TFC7\left({{I_{i,k}}}\right)\right)$. The training process of network is same as above baseline as shown in Section \ref{sec:base_l}. So a series of trained CNN models are obtained for corresponding to each part subset.

In the Testing Phase, first, the pedestrian images are also divided into several parts as same as the samples of training set.

Then, for the parts of new query and testing pedestrian image, we calculate the binary feature with the learned parameters of each layers. For the $k$-th part of pedestrian image $I_j$, the hash code is calculated as follows:
\begin{equation}
\hat{H}\left( {{I_{j,k}}} \right) = {sign}\left({f_H}\left( {{I_{i,k}}} \right) - 0.5 \right).
\end{equation}
In this way, a group of hash codes is obtained for any parts of a single pedestrian image.

Finally, the hash codes of query and testing image $I_j$ is represented by concatenating each part.
\begin{equation}
\hat{H}\left( {{I_j}} \right) = {{\rm concatenation}}\{ \hat{H}\left( {{I_{j,k}}} \right)\} ,k = 1,...,M.\end{equation}
In this way, we finish a hash codes conversion of local parts to global image. The new part-based hash codes can extract some rich and useful descriptors that retain the spatial information.

After the generation of hash codes for query and testing pedestrian image dataset, the person re-id is evaluated by calculating and sorting the Hamming distance between query and testing samples.

\subsection{Multiple Queries}
The motivation of multiple queries is that the intra-class variation of samples is taken into consideration. The strategy of multiple queries is to merge the query images which belong to same identity under a single camera into a single query for speed consideration. The method of multiple queries, which is more robust to pedestrian variations, has shown superior performance in image search \cite{arandjelovic2012multiple} and person re-id \cite{farenzena2010person}. We implement two pooling strategies, which are average pooling and max pooling, respectively. In average pooling, the feature vectors of multiple queries are pooled into one vector by averaged sum. In max pooling, the feature vectors of multiple queries are pooled into one vector by taking the maximum value in each dimension from all queries.

\setlength{\tabcolsep}{10.4pt}
\begin{table}[t]
\centering
\caption{Baseline Performance (Rank-1 Accuracy (\%) and mAP (\%)) with Different Hash Codes Lengths on Market-1501 and Market-1501+500K Datasets.}
\begin{tabular}{l|cc|cc} \hline
\multirow{2}{*}{Hash Codes Length} & \multicolumn{2}{c}{Market-1501} & \multicolumn{2}{|c}{Market-1501+500K}\\ \cline{2-5}
& r=1 & mAP & r=1 & mAP\\ \hline
128 bits  & 19.06 & 9.58 & 12.20 & 4.48\\
256 bits  & 21.91 & 10.75 & 14.82 & 5.33\\
512 bits  & 25.36 & 12.37 & 17.61 & 6.44\\
1,024 bits  & 25.24 & 11.95 & 16.63 & 5.91\\
2,048 bits  & 27.14 & 12.76 & 18.68 & 6.56\\ \hline
\end{tabular}
\label{table:baseline}
\end{table}

\setlength{\tabcolsep}{10.4pt}
\begin{table*}[t]
\centering
\caption{Rank-1 Accuracy (\%) and mAP (\%) for Part-based Model on Market-1501 and Market-1501+500K. ``MQ'' Represents Multiple Queries. The ``avg'' and ``max'' Denote Average and Max Pooling, Respectively.}
\begin{tabular}{l|cc|cc|cc|cc|cc|cc} \hline
\multirow{3}{*}{Methods} & \multicolumn{6}{c|}{Market-1501}& \multicolumn{6}{c}{Market-1501+500K}\\ \cline{2-13}
& \multicolumn{2}{c|}{Single Query} & \multicolumn{2}{c|}{MQ avg} & \multicolumn{2}{c|}{MQ max} & \multicolumn{2}{c|}{Single Query} & \multicolumn{2}{c|}{MQ avg} & \multicolumn{2}{c}{MQ max}\\\cline{2-13}
& r=1& mAP& r=1& mAP& r=1& mAP& r=1& mAP& r=1& mAP& r=1& mAP\\ \hline
Entire & 27.14 & 12.76& 32.54 & 15.39 &30.17& 14.66 & 18.68 & 6.56& 21.91 & 7.82 & 19.69 & 7.31\\
3 Parts & 43.05 & 21.80& 49.52 & 26.89& 46.70 & 25.25 & 32.28 & 13.21 & 39.34 & 17.26 & 36.46 & 15.72\\
4 Parts & \textbf{47.24} & \textbf{24.94}& \textbf{57.13} & \textbf{31.03}& \textbf{54.39} & \textbf{29.74} & \textbf{37.23} & \textbf{16.38} & \textbf{45.72} & \textbf{21.26} & \textbf{43.53} & \textbf{19.93}\\
5 Parts & 46.23 & 23.72& 53.86 & 29.46& 51.07 & 28.12 & 37.08 & 15.44 & 44.63 & 20.61 & 41.48 & 19.10\\ \hline
\end{tabular}\label{table:MQ}
\end{table*}

\setlength{\tabcolsep}{9.4pt}
\begin{table*}[t]
\centering
\caption{Rank-1 Accuracy (\%) and mAP (\%) for Different Region Partitioning Strategies. ``EQL'' and ``UnEQL'' Denote Partitioning Images Equally and Unequally, Respectively.}
\begin{tabular}{l|cc|cc|cc|cc|cc|cc} \hline
\multirow{3}{*}{Partitions} & \multicolumn{6}{c|}{Market-1501} & \multicolumn{6}{c}{Market-1501+500K}\\ \cline{2-13}
& \multicolumn{2}{c|}{Single Query} & \multicolumn{2}{c|}{MQ avg} & \multicolumn{2}{c|}{MQ max} & \multicolumn{2}{c|}{Single Query} & \multicolumn{2}{c|}{MQ avg} & \multicolumn{2}{c}{MQ max}\\ \cline{2-13}
& r=1& mAP& r=1& mAP& r=1& mAP& r=1& mAP& r=1& mAP& r=1& mAP\\ \hline
EQL 3 Parts & 43.05 & 21.80 & 49.52 & 26.89 & 46.70 & 25.25 & 32.28 & 13.21 & 39.34 & 17.26 & 36.46 & 15.72\\
UnEQL 3 Parts & 36.72 & 18.56 & 46.41 & 24.18 & 43.08 & 22.38 & 26.34 & 10.69 & 36.34 & 15.36 & 31.21 & 13.38\\
Overlap 3 Parts & 47.36 & 25.47 & 53.36 & 30.29 & 50.86 & 28.54 & 37.47 & 16.19 & 42.70 & 19.96 & 39.61 & 18.40\\
EQL 4 Parts & 47.24 & 24.94 & \textbf{57.13} & 31.03 & \textbf{54.39} & 29.74 & 37.23 & 16.38 & \textbf{45.72} & 21.26 & \textbf{43.53} & 19.93\\
UnEQL 4 Parts & 46.17 &24.31 & 54.69 &30.48 & 51.57 & 29.16 & 35.99 & 15.47 & 44.80 & \textbf{21.29} & 42.04 & \textbf{20.02}\\
Overlap 4 Parts & \textbf{47.89} & \textbf{26.06} & 56.80 &\textbf{31.67} & 53.83 & \textbf{30.40} & \textbf{38.39} & \textbf{16.82} & 45.64 & 21.16 & 41.83 & 19.99\\ \hline
\end{tabular}\label{table:partition}
\end{table*}

\section{Experiments}\label{sect:4}
In this section, we first describe the datasets and evaluation protocol. Then we evaluate the proposed PDH method and provide some comparisons with the state-of-the-art hashing and person re-id methods to demonstrate the effectiveness and efficiency of the PDH method.

\subsection{Datasets and Evaluation Protocol}
This paper evaluates the performance of the proposed PDH method on the largest person re-id dataset: Market-1501  \cite{zheng2015scalable} and its associating distractor set with 500K images. The two datasets are denoted as: \textbf{Market-1501} and \textbf{Market-1501+500K}, respectively. The Market-1501 dataset contains 32,668 bounding boxes of 1,501 identities. There are 14.8 cross-camera ground truths for each query on average. The testing process is performed in a cross-camera mode. The distractor set contains 500K images which are treated as outliers besides the 32,668 bounding boxes of 1,501 identities. The Market-1501 is currently the largest person re-id dataset  which is closer towards realistic situations than previous ones. We choose these two datasets due to their scales,  for which effective retrieval methods are of great needs.

\setlength{\tabcolsep}{9.4pt}
\begin{table*}[t]
\centering
\caption{Rank-1 Accuracy (\%) and mAP (\%) of Each Individual Part and Concatenation for Different Region Partitioning Strategies. ``EQL'' and ``UnEQL'' Denote Partitioning Images Equally and Unequally, Respectively.}
\begin{tabular}{l|cc|cc|cc|cc|cc|cc} \hline
\multirow{3}{*}{Methods} & \multicolumn{6}{c|}{Market-1501} & \multicolumn{6}{c}{Market-1501+500K}\\ \cline{2-13}
& \multicolumn{2}{c|}{EQL 4 Parts} & \multicolumn{2}{c|}{UnEQL 4 Parts}& \multicolumn{2}{c|}{Overlap 4 Parts} & \multicolumn{2}{c|}{EQL 4 Parts} & \multicolumn{2}{c|}{UnEQL 4 Parts}& \multicolumn{2}{c}{Overlap 4 Parts} \\\cline{2-13}
& r=1 & mAP & r=1 & mAP & r=1 & mAP & r=1 & mAP & r=1 & mAP & r=1 & mAP \\ \hline
CNN-1 (Part 1) & 6.83 & 3.21 & 5.82 & 2.75 & 11.43 & 4.89 & 3.33 & 1.07 & 2.58 & 0.87 & 6.65 & 1.81 \\
CNN-2 (Part 2) & 11.49 & 4.83 & 14.55 & 5.98 & 19.12 & 8.35 & 6.41 & 1.77 & 8.70 & 2.43 & 11.64 & 3.71 \\
CNN-3 (Part 3) & 10.66 & 5.13 & 7.54 & 3.59 & 19.69 & 9.15 & 5.85 & 1.97 & 4.04 & 1.33 & 11.52 & 4.06 \\
CNN-4 (Part 4) & 3.36 & 1.45 & 1.93 & 0.99 & 5.70 & 2.76 & 1.84 & 0.44 & 1.01 & 0.24 & 3.15 & 0.97 \\\hline
\bf Concatenation & \bf 47.24 & \bf 24.94 & \bf 46.17 & \bf 24.31 & \bf 47.89 & \bf 26.06 & \bf 37.23 & \bf 16.38 & \bf 35.99 & \bf 15.47 & \bf 38.39 & \bf 16.82 \\ \hline
\end{tabular}\label{table:pingfen}
\end{table*}

We adopt the Cumulated Matching Characteristics (CMC) curve and mean Average Precision (mAP) on Market-1501 and Market-1501+500K datasets. The CMC curve shows the probability that a query identity appears in the ranking lists of different sizes. The rank-1 accuracy (r=1) is shown when CMC curves are absent. The CMC is generally believed to focus on precision. In case of there is only one ground truth match for a given query, the precision and recall are the same. However, if multiple ground truths exist, the CMC curve is biased because recall is not considered. For Market-1501 and Market-1501+500K datasets, there are several cross-camera ground truths for each query. The mean Average Precision (mAP) is more suitable to evaluate the overall performance. The mAP considers both the precision and recall, thus providing a more comprehensive evaluation.

\subsection{Experimental Results}

\subsubsection{Performance of the Baseline Method}\label{sec:ba}
We evaluate the baseline deep hashing model (described in Section \ref{sec:base_l}) trained by the entire pedestrian images. We observe from Table \ref{table:baseline} that the baseline produces a relatively low accuracy on Market-1501 and Market-1501+500K datasets. Hash codes with various lengths are tested on the two datasets. It is shown from the results that longer hash codes generally yield higher re-id accuracy. The increase is most evident for shorter hash codes. For hash codes of more than 512 bits, re-id accuracy remains stable or witnesses some slight decrease. \textbf{As a trade-off between efficiency and accuracy, we use the 512 bits hash codes for each part-based deep CNN model in the following experiments.}

\subsubsection{Impact of Part Integration}
In Table \ref{table:MQ}, we evaluate the impact of part integration on re-id accuracy, with a comparison with the baseline on the two re-id datasets. The entire pedestrian image is partitioned into several equal parts horizontally. We observe from Table \ref{table:MQ} that when partitioning into 4 parts, mAP increases from 12.76\% to 24.94\% (+12.18\%), and an even larger improvement can be seen from rank-1 accuracy, from 27.14\% to 47.24\% (+20.10\%) on Market-1501 dataset. On Market-1501+500K dataset, mAP increases from 6.56\% to 16.38\% (+9.82\%) with 4 parts, and for rank-1 accuracy, from 18.68\% to 37.23\% (+18.55\%). This illustrates the effectiveness of the part integration over the baseline method. Moreover, we find that using more parts typically produces higher re-id performance, but again, the improvement tends to saturate after 4 parts.

We then evaluate multiple queries on the two re-id datasets. The experimental results demonstrate that the usage of multiple queries improves 4\%$\sim$7\% in mAP and 3\%$\sim$10\% in rank-1 accuracy. Moreover, multiple queries by average pooling is slightly superior to max pooling. The performance of part-based model increases significantly compared with the original general deep hashing model. These results demonstrate the effectiveness of part-based model and multiple queries for large-scale person re-id.

\subsubsection{Comparison of Different Part Partitioning Strategies}
The Section \ref{sec:part_model} describes 6 part partitioning variants. Specifically, the three types of part partitioning strategies are evaluated, including ``Equally'', ``Unequally'' and ``Overlap''. The height of the original  pedestrian image is $128$. The direction of region partition is along with horizontal. The partition details are listed in Table \ref{table:part_size}.

In Table \ref{table:partition}, we provide a comparison among these partitioning strategies. Results suggest that generating parts with overlap is an effective way of training the CNN model, probably because the overlaps provide some complementary information between two adjacent parts. We observe from Table \ref{table:partition} that when using ``Overlap 4 parts'', rank-1 accuracy increases from 47.24\% to 47.89\% (+0.65\%), and an even larger improvement can be seen from mAP, from 24.94\% to 26.06\% (+1.12\%) on Market-1501 dataset. On Market-1501+500K dataset, rank-1 accuracy increases from 16.38\% to 16.82\% (+0.44\%) with 4 parts, and for mAP, from 37.23\% to 38.39\% (+1.16\%).
Meanwhile, the unequal part partition is inferior to equal parts, especially on the results of 3 parts. We speculate that the non-uniform operation separates some parts which have specific semantic meanings.

\setlength{\tabcolsep}{9.4pt}
\begin{table*}[t]
\centering
\caption{Rank-1 Accuracy (\%) and mAP (\%) Comparing whether to Share Weights among the Part Sub-networks. ``EQL'' and ``UnEQL'' Denote Partitioning Images Equally and Unequally, Respectively.}
\begin{tabular}{l|cc|cc|cc|cc|cc|cc} \hline
\multirow{3}{*}{Methods} & \multicolumn{6}{c|}{Market-1501} & \multicolumn{6}{c}{Market-1501+500K}\\ \cline{2-13}
& \multicolumn{2}{c|}{EQL 4 Parts} & \multicolumn{2}{c|}{UnEQL 4 Parts} & \multicolumn{2}{c|}{Overlap 4 Parts} & \multicolumn{2}{c|}{EQL 4 Parts} & \multicolumn{2}{c|}{UnEQL 4 Parts} & \multicolumn{2}{c}{Overlap 4 Parts}\\ \cline{2-13}
& r=1& mAP& r=1& mAP& r=1& mAP& r=1& mAP& r=1& mAP& r=1& mAP\\ \hline
share weights & 41.42 & 19.57 & 33.97 & 16.13 & 42.96 & 21.70 & 32.39 & 12.40 & 25.21 & 9.36 & 34.09 & 13.74 \\ \hline
not share weights & \textbf{47.24} & \textbf{24.94} & \textbf{46.17} & \textbf{24.31} & \textbf{47.89} & \textbf{26.06} & \textbf{37.23} & \textbf{16.38} & \textbf{35.99} & \textbf{15.47} & \textbf{38.39} & \textbf{16.82} \\ \hline
\end{tabular}\label{table:part_share_compare}
\end{table*}

\setlength{\tabcolsep}{12.4pt}
\begin{table*}[t]
\centering
\caption{Rank-1 Accuracy (\%) and mAP (\%) Comparison with the State-of-the-Art Hashing Methods.}
\begin{tabular}{l|cc|cc|cc|cc|cc} \hline
\multirow{3}{*}{Methods} & \multicolumn{10}{c}{Market-1501}\\ \cline{2-11}
& \multicolumn{2}{c|}{128 bits} & \multicolumn{2}{c|}{256 bits} & \multicolumn{2}{c|}{512 bits} & \multicolumn{2}{c|}{1,024 bits} & \multicolumn{2}{c}{2,048 bits}\\\cline{2-11}
& r=1 & mAP & r=1 & mAP & r=1 & mAP & r=1 & mAP & r=1 & mAP\\ \hline
SH-CNN \cite{Weiss2008Spectral} & 34.35 & 16.26 & 36.08 & 19.50 & 39.86 & 21.75 & 42.65 & 23.28 & 44.23 & 23.73\\
USPLH-CNN \cite{Wang2010Sequential} & 33.31 & 16.42 & 37.62 & 18.35 & 38.42 & 18.92 & 39.04 & 18.84 & 39.93 & 18.82\\
SpH-CNN \cite{Heo2012Spherical} & 35.33 & 16.65 & 37.96 & 19.99 & 41.38 & 22.56 & 44.26 & 24.04 & 44.62 & 24.54\\
DSH-CNN \cite{Jin2014Density} & 33.11 & 16.17 & 38.09 & 19.21 & 41.95 & 21.22 & 43.50 & 22.15 & 44.86 & 22.70\\
KSH-CNN \cite{Liu2012Supervised} & 41.33 & 20.62 & 43.55 & 23.40 & 44.23 & 24.41 & 45.27 & 24.90 & 46.13 & 25.01\\
SDH-CNN \cite{shen2015supervised} & 35.69 & 17.82 & 39.01 & 20.59 & 40.41 & 21.93 & 41.81 & 22.28 & 43.56 & 23.19\\ \hline
Zhang \emph{et al.} \cite{zhang2015bit} & 15.50 & 8.50 & 18.38 & 9.48 & 22.24 & 11.07 & 21.91 & 10.47 & 23.43 & 11.29\\
Lin \emph{et al.} \cite{lin2015deep} & 8.91 & 4.89 & 18.65 & 10.01 & 28.98 & 16.39 & 41.12 & 24.14 & 49.79 & 30.29\\ \hline
\textbf{Our PDH method} & 36.31 & 19.59 & 42.07 & 22.43 & 44.60 & 24.30 & 49.58 & 26.09 & 47.89 & 26.06\\ \hline
\end{tabular}
\begin{tabular}{l|cc|cc|cc|cc|cc} \hline
\multirow{3}{*}{Methods} & \multicolumn{10}{c}{Market-1501+500K}\\ \cline{2-11}
& \multicolumn{2}{c|}{128 bits} & \multicolumn{2}{c|}{256 bits} & \multicolumn{2}{c|}{512 bits} & \multicolumn{2}{c|}{1,024 bits} & \multicolumn{2}{c}{2,048 bits}\\\cline{2-11}
& r=1 & mAP & r=1 & mAP & r=1 & mAP & r=1 & mAP & r=1 & mAP\\ \hline
SH-CNN \cite{Weiss2008Spectral} & 14.05 & 6.03 & 16.98 & 7.30 & 22.26 & 9.05 & 25.38 & 10.51 & 28.89 & 11.48\\
USPLH-CNN \cite{Wang2010Sequential} & 12.56 & 4.91 & 15.32 & 6.03 & 15.80 & 6.46 & 16.48 & 6.59 & 17.19 & 6.71\\
SpH-CNN \cite{Heo2012Spherical} & 17.96 & 6.80 & 23.40 & 9.17 & 26.51 & 10.65 & 28.30 & 11.60 & 29.60 & 12.08\\
DSH-CNN \cite{Jin2014Density} & 10.51 & 3.40 & 14.16 & 4.92 & 14.96 & 5.32 & 15.02 & 5.17 & 18.53 & 6.25\\
KSH-CNN \cite{Liu2012Supervised} & 30.94 & 11.86 & 35.39 & 13.99 & 36.28 & 14.88 & 37.62 & 15.17 & 37.44 &  15.37\\
SDH-CNN \cite{shen2015supervised} & 23.75 & 8.61 & 26.75 & 10.34 & 29.75 & 11.72 & 31.86 & 12.45 & 30.79 & 12.25\\ \hline
Zhang \emph{et al.} \cite{zhang2015bit} & 9.71 & 3.65 & 11.19 & 4.25 & 14.49 & 5.20 & 13.66 & 4.70 & 14.52 & 5.24\\
Lin \emph{et al.} \cite{lin2015deep} & 5.34 & 1.99 & 10.90 & 4.41 & 18.85 & 7.83 & 28.92 & 13.15 & 37.41 & 18.26\\ \hline
\textbf{Our PDH method} & 27.05 & 11.58 & 31.80 & 13.43 & 34.17 & 15.04 & 39.34 & 16.77 & 38.39 & 16.82\\ \hline
\end{tabular}\label{table:compare_hash}
\end{table*}

In order to further investigate the role of different individual parts, we evaluate the re-id performance of individual parts, and compare it with the concatenation of all parts. The hash code of each part is generated by the training CNN model at the corresponding regions. We observe from Table \ref{table:pingfen} that each individual CNN model produces a low accuracy on Market-1501 and Market-1501+500K datasets, especially the CNN-1 and CNN-4 models. However, after the concatenation of hash codes for all the parts, the re-id accuracy is improved dramatically. The experimental results thus demonstrate that the part partitioning is effective in the proposed method.

The sub-networks proposed in this paper do not share weights, because the body parts are different in nature. To illustrate this point, we conduct experiments comparing whether to share weights among the part sub-networks. We train the CNN models using the weight-sharing network for the parts and provide some experimental results and comparisons in Table \ref{table:part_share_compare}. It can be observed that the accuracy of weight-sharing network can be over 6\% lower than training the sub-networks independently, which validating our assumption.

\subsubsection{Comparison with the State-of-the-art Hashing Methods}
In this section, we compare the proposed PDH method with some state-of-the-art hashing methods on Market-1501 and Market-1501+500K datasets. The compared hashing methods include Spectral Hashing (SH) \cite{Weiss2008Spectral}, Unsupervised Sequential Projection Learning Hashing (USPLH) \cite{Wang2010Sequential}, Spherical Hashing (SpH) \cite{Heo2012Spherical}, Density Sensitive Hashing (DSH) \cite{Jin2014Density}, Kernel Supervised Hashing (KSH) \cite{Liu2012Supervised}, Supervised Discrete Hashing (SDH) \cite{shen2015supervised} and two deep hashing methods \cite{zhang2015bit}, \cite{lin2015deep}. The first four methods are unsupervised and the others are supervised hashing methods. For the two comparison deep hashing methods \cite{zhang2015bit} and \cite{lin2015deep}, which are Siamese and identification CNN model, respectively. We use the image pixels as input directly and implement it based on the Caffe \cite{Jia2014Caffe} framework for deep hashing. The conventional non-deep hashing methods are evaluated based on the 4,096-D FC7 features in CaffeNet \cite{krizhevsky2012imagenet} pre-trained on the ImageNet \cite{deng2009imagenet} dataset and fine-tuned on the training set of the Market-1501 dataset for fair comparison. This feature is also called ID-discriminative Embedding (IDE) in \cite{zheng2016mars}.

\setlength{\tabcolsep}{9.4pt}
\begin{table*}[t]
\centering
\caption{Rank-1 Accuracy (\%) and mAP (\%) Comparison with Other Deep Hashing Methods for Different Region Partitioning Strategies. ``EQL'' and ``UnEQL'' Denote Partitioning Images Equally and Unequally, Respectively.}
\begin{tabular}{l|cc|cc|cc|cc|cc|cc} \hline
\multirow{3}{*}{Methods} & \multicolumn{6}{c|}{Market-1501} & \multicolumn{6}{c}{Market-1501+500K}\\ \cline{2-13}
& \multicolumn{2}{c|}{EQL 4 Parts} & \multicolumn{2}{c|}{UnEQL 4 Parts} & \multicolumn{2}{c|}{Overlap 4 Parts} & \multicolumn{2}{c|}{EQL 4 Parts} & \multicolumn{2}{c|}{UnEQL 4 Parts} & \multicolumn{2}{c}{Overlap 4 Parts}\\ \cline{2-13}
& r=1& mAP& r=1& mAP& r=1& mAP& r=1& mAP& r=1& mAP& r=1& mAP\\ \hline
Zhang \emph{et al.} \cite{zhang2015bit} & 38.90 & 20.14 & 40.08 & 19.86 & 41.63 & 21.91 & 29.78 & 12.17 & 29.45 & 11.54 & 30.73 & 13.05 \\
Lin \emph{et al.} \cite{lin2015deep} & 48.60 & 26.82 & 41.89 & 22.25 & 49.55 & 28.25 & 36.37 & 16.53 & 29.48 & 12.79 & 37.35 & 17.56 \\ \hline
\bf Our PDH Method & 47.24 & 24.94 & 46.17 & 24.31 & 47.89 & 26.06 & 37.23 & 16.38 & 35.99 & 15.47 & 38.39 & 16.82 \\ \hline
\end{tabular}\label{table:part_deep_compare}
\end{table*}

\begin{figure*} [t]
\centering
\subfigure[Market-1501]{\label{fig:cmc_market}
\includegraphics[width=0.48\textwidth{}]{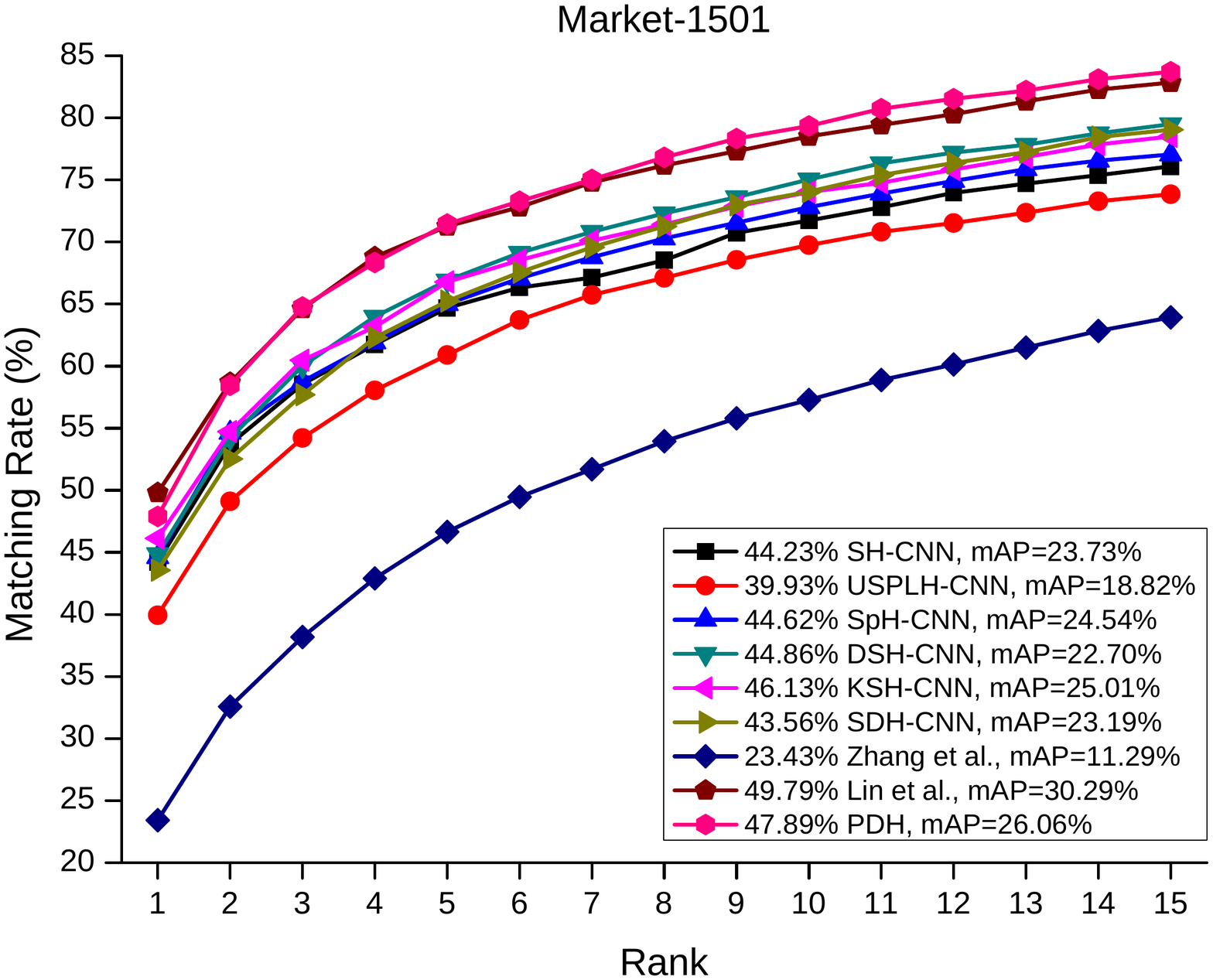}}
\hspace{0.02in}
 \subfigure[Market-1501+500K]{\label{fig:cmc_500k}
\includegraphics[width=0.48\textwidth{}]{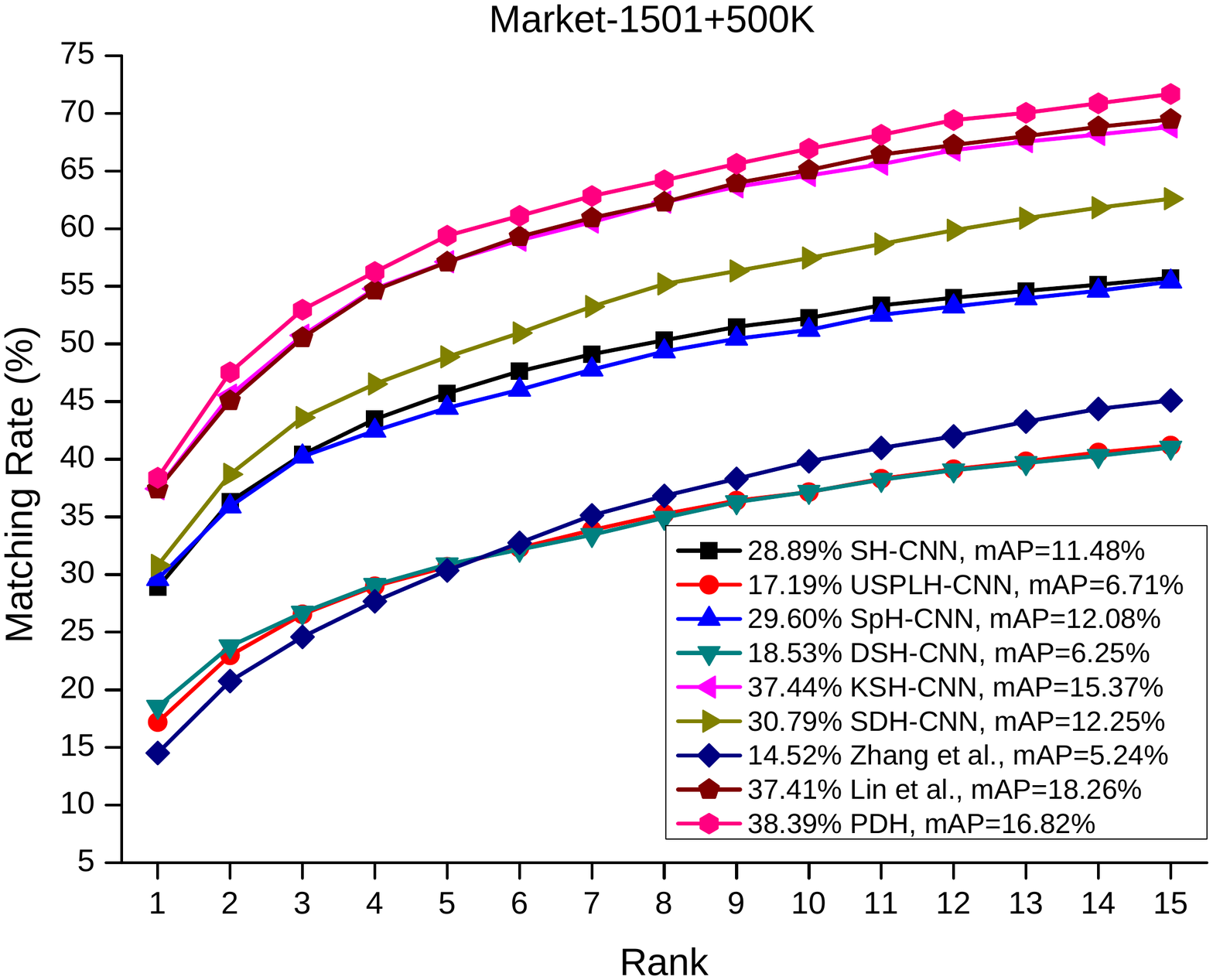}}
\caption{CMC curves of the state-of-the-art hashing methods on Market-1501 and Market-1501+500K datasets.}
\label{fig:cmc}
\end {figure*}

Table \ref{table:compare_hash} summarizes the results of the state-of-the-art hashing methods with different code lengths on Market-1501 and Market-1501+500K datasets. Fig. \ref{fig:cmc} shows the CMC curve comparison of different hashing methods at 2,048 bits code length. First, it is evident that when longer hash codes are used, the rank-1 accuracy and mAP increase significantly. Second, compared with unsupervised hashing methods, the conventional non-deep supervised hashing methods (KSH and SDH) generally achieve better performance. Third, the bit-scalable deep hashing method \cite{zhang2015bit} produces a relatively low accuracy, similar to the baseline in Section \ref{sec:ba}. For \cite{zhang2015bit}, using 1,024 bits is inferior to using 512 bits in both rank-1 accuracy and mAP as shown in Table \ref{table:compare_hash}. We also notice a similar trend for the baseline method in Table \ref{table:baseline}. In fact, it is common that the retrieval accuracy becomes saturated as the hash code grows longer, so after 512 bits, there might be some small fluctuations in the accuracy. Fourth, comparing with \cite{lin2015deep}, we show that \cite{lin2015deep} produces a superior mAP at 2,048 bits. However, the rank-1 accuracy and mAP of \cite{lin2015deep} decline significantly with the decrease of the hash codes length, and is inferior to our PDH method in these cases.

Compared with these hashing methods, the proposed PDH method produces a competitive performance \emph{w.r.t.} rank-1 accuracy, mAP, and the CMC curve when 2,048 bits are used. Specifically, our method achieves \textbf{rank-1 accuracy = 47.89\% and mAP = 26.06\% on Market-1501 dataset, rank-1 accuracy = 38.39\% and mAP = 16.82\% on Market-1501+500K dataset,} respectively.

In order to further study the scalability of our part integration, we evaluate the part integration on above two deep hashing methods \cite{zhang2015bit}, \cite{lin2015deep} with a comparison of the proposed PDH method. We use the 512 bits hash vectors for each part-based deep CNN model. From the results in Table \ref{table:part_deep_compare}, the accuracy of part integration on two deep hashing methods \cite{zhang2015bit}, \cite{lin2015deep} have increased their baselines (as shown in Table \ref{table:compare_hash}) by a large margin.

The experimental results demonstrate that our PDH method produce a competitive performance for large-scale person re-id. Moreover, the part integration has superior scalability on other deep hashing methods.

\subsubsection{Comparison with the State-of-the-art Person Re-id Methods}
We first compare with the Bag-of-Words (BOW) descriptor \cite{zheng2015scalable}. We only list the best result in \cite{zheng2015scalable}. As can be seen in Table \ref{table:compare_reid}, the proposed PDH method brings decent improvement of benchmark in both rank-1 accuracy and mAP. In addition, we compare with some existing metric learning methods based on BOW descriptor. The metric learning methods include LMNN \cite{weinberger2005distance}, ITML \cite{davis2007information} and KISSME \cite{koestinger2012large}. From the results in Table \ref{table:compare_reid}, it is clear that the proposed PDH method significantly outperforms the traditional pipeline approaches, which demonstrates the effectiveness of proposed PDH method.

Then we compare with some state-of-the-art person re-id methods based on deep learning, including Multi-region Bilinear Convolutional Neural Networks method  \cite{ustinova2015multiregion}, PersonNet method \cite{wu2016personnet}, Semi-supervised Deep Attribute Learning (SSDAL) method \cite{Su2016Deep}, Temporal Model Adaptation (TMA) method \cite{martinel2016temporal} and End-to-end Comparative Attention Network (CAN) method \cite{liu2016end}. From the results in Table \ref{table:compare_reid}, it is clear that the proposed PDH method significantly outperforms most of deep learning based re-id methods in both rank-1 accuracy and mAP. Only the PDH (MQ avg) is slightly inferior to Multi-region Bilinear DML (MQ avg) \cite{ustinova2015multiregion} in mAP. Nevertheless, the advantage of our method lies in the binary signatures, which enable fast person re-id in large galleries. In summary, PDH yields competitive accuracy on Market-1501, but has the advantage of computational and storage efficiency.
\makeatother

\subsubsection{Comparison of Total Coding Time with Different Person Re-id Methods During the Testing Phase} We compare the total coding time of PDH with two existing methods, including the 5,600-D BOW descriptor based on Color Names \cite{zheng2015scalable} and the 4,096-D IDE descriptor \cite{zheng2016mars} (FC7 features in CaffeNet \cite{krizhevsky2012imagenet} pretrained on the ImageNet \cite{deng2009imagenet} dataset and fine-tuned on the
training set of Market-1501 dataset). The coding time during the testing phase is composed of three aspects: 1) feature extraction, 2) average search (distance calculation), and 3) sorting. On the one hand, the computation of Hamming distance is much faster than Euclidean distance. On the other hand, using the Bucket sorting algorithm, the sorting complexity of PDH is $O(n)$, which is much lower than the baseline sorting complexity $O(n*\log (n))$ for floating-point vectors.

\setlength{\tabcolsep}{11.7pt}
\begin{table}
\centering
\caption{Rank-1 Accuracy (\%) and mAP (\%) Comparison with the State-of-the-Art Person Re-id Methods. Note that We only Compare with the ``OLDER'' Methods as this work was done in late 2015.}
\begin{tabular}{l|cc} \hline
\multirow{2}{*}{Methods} & \multicolumn{2}{c}{Market-1501}\\ \cline{2-3}
& r=1& mAP\\ \hline
BOW+HS \cite{zheng2015scalable} & 47.25 &21.88\\
BOW+LMNN \cite{weinberger2005distance} & 34.00 &15.66\\
BOW+ITML \cite{davis2007information} & 38.21 & 17.05\\
BOW+KISSME \cite{koestinger2012large} & 39.61 & 17.73\\
Bit-scalable Deep Hashing \cite{zhang2015bit} & 23.43 & 11.29\\
Multiregion Bilinear (Single Query) \cite{ustinova2015multiregion} & 45.58 & 26.11\\
Multiregion Bilinear (MQ avg) \cite{ustinova2015multiregion} & 56.59 & \bf32.26\\
Multiregion Bilinear (MQ max) \cite{ustinova2015multiregion} & 53.62 & 30.76\\
PersonNet (Single Query)  \cite{wu2016personnet} & 37.21 & 18.57\\
SSDAL (Single Query) \cite{Su2016Deep} & 39.40 & 19.60\\
SSDAL (MQ avg) \cite{Su2016Deep} & 48.10 & 25.40\\
SSDAL (MQ max) \cite{Su2016Deep} & 49.00 & 25.80\\
TMA (Single Query) \cite{martinel2016temporal} & 47.92 & 22.31\\
End-to-end CAN (Single Query) \cite{liu2016end} & 48.24  & 24.43\\
\hline
Our PDH (Single Query) & 47.89 & 26.06\\
Our PDH (MQ max) & 53.83 & 30.40\\
\textbf{Our PDH (MQ avg)} & \bf 56.80 &  31.67\\ \hline
\end{tabular}\label{table:compare_reid}
\end{table}

\setlength{\tabcolsep}{6.4pt}
\begin{table*}
\centering
\caption{Feature Extraction, Distance Calculation, Sorting and Total Coding Time of the IDE, BOW Features and the Proposed PDH on Market-1501 and Market-1501+500K (Partition by ``/'', Millisecond Per Image) Datasets.}
\begin{tabular}{l|c|c|ccc|c}
\hline Methods & Dim.&Data Type&Feature Extraction &Distance Calculation&Sorting&Total Coding Time\\
\hline IDE (FC7) \cite{zheng2016mars} &4,096& Float&8.3/8.3 &97.9/2,470.8&3.5/134.5&109.7/2,613.6\\
 BOW (CN) \cite{zheng2015scalable} &5,600 &Float&264.3/264.3 & 139.9/3,587.9&4.9/156.1&409.1/4,008.3\\
 \hline
\textbf{Our PDH method}&2,048& Bool& 32.8/32.8& 0.98/26.2&0.83/16.8&\bf 34.61/75.8\\
\hline
\end{tabular}\label{table:timing}
\end{table*}

Table \ref{table:timing} presents the feature extraction, distance calculation, sorting and total coding time (millisecond (ms)) of the three methods on Market-1501 and Market-1501+500K datasets. The evaluation is performed on a server with GTX 1080 GPU (8G memory), 2.60 GHz CPU and 128 GB memory. The feature extraction time of the proposed PDH method is 32.8 ms, which is slower than IDE features due to the multiple parts evaluation. However, in practice, we can extract the features of each part for an image in parallel, and accelerate the feature extraction process of PDH method. Therefore, the disadvantage in the feature extraction time could be reduced. The search time of the PDH method is 0.98 ms and 26.2 ms on the two re-id datasets, respectively. While, the sorting time of the PDH method is 0.83 ms and 16.8 ms, respectively, which is much faster than the other two float-point feature representations. From the total coding time comparison of three methods, the efficiency of the proposed PDH method should be justified. With the growth of the scale of person re-id datasets, binary representations will become more important.

\section{Conclusions}\label{sect:5}
In this paper, we employ the triplet-based deep hashing model and propose a \textbf{P}art-based \textbf{D}eep \textbf{H}ashing (PDH) framework for improving the efficiency and accuracy of large-scale person re-id, which generates hash codes for pedestrian images via a well designed part-based deep architecture. The part-based representation increases the discriminative ability of visual matching, and provides a significant improvement over the baseline. Multiple queries method is rewarding to improve the person re-id performance. The proposed PDH method demonstrates very competitive performance compared with state-of-the-art re-id methods on large-scale Market-1501 and Market-1501+500K datasets. There are several challenging directions along which we will extend this work. First, larger databases with millions of bounding boxes will be built which will fully show the strength of hashing methods. Second, more discriminative CNN models will be investigated to learn effective binary representations.

\ifCLASSOPTIONcaptionsoff
  \newpage
\fi

\bibliographystyle{IEEEtran}
\bibliography{mybibtxt}

\begin{thebibliography}{10}
\providecommand{\url}[1]{#1}
\csname url@samestyle\endcsname
\providecommand{\newblock}{\relax}
\providecommand{\bibinfo}[2]{#2}
\providecommand{\BIBentrySTDinterwordspacing}{\spaceskip=0pt\relax}
\providecommand{\BIBentryALTinterwordstretchfactor}{4}
\providecommand{\BIBentryALTinterwordspacing}{\spaceskip=\fontdimen2\font plus
\BIBentryALTinterwordstretchfactor\fontdimen3\font minus
  \fontdimen4\font\relax}
\providecommand{\BIBforeignlanguage}[2]{{%
\expandafter\ifx\csname l@#1\endcsname\relax
\typeout{** WARNING: IEEEtran.bst: No hyphenation pattern has been}%
\typeout{** loaded for the language `#1'. Using the pattern for}%
\typeout{** the default language instead.}%
\else
\language=\csname l@#1\endcsname
\fi
#2}}
\providecommand{\BIBdecl}{\relax}
\BIBdecl

\bibitem{yan2016image}
Y.~Yan, F.~Nie, W.~Li, C.~Gao, Y.~Yang, and D.~Xu, ``Image classification by
  cross-media active learning with privileged information,'' \emph{IEEE
  Transactions on Multimedia}, vol.~18, no.~12, pp. 2494--2502, 2016.

\bibitem{yang2013feature}
Y.~Yang, Z.~Ma, A.~G. Hauptmann, and N.~Sebe, ``Feature selection for
  multimedia analysis by sharing information among multiple tasks,'' \emph{IEEE
  Transactions on Multimedia}, vol.~15, no.~3, pp. 661--669, 2013.

\bibitem{chang2016compound}
X.~Chang, F.~Nie, S.~Wang, Y.~Yang, X.~Zhou, and C.~Zhang, ``Compound rank-$ k
  $ projections for bilinear analysis,'' \emph{IEEE Transactions on Neural
  Networks and Learning Systems}, vol.~27, no.~7, pp. 1502--1513, 2016.

\bibitem{yang2008harmonizing}
Y.~Yang, Y.-T. Zhuang, F.~Wu, and Y.-H. Pan, ``Harmonizing hierarchical
  manifolds for multimedia document semantics understanding and cross-media
  retrieval,'' \emph{IEEE Transactions on Multimedia}, vol.~10, no.~3, pp.
  437--446, 2008.

\bibitem{yang2012multimedia}
Y.~Yang, F.~Nie, D.~Xu, J.~Luo, Y.~Zhuang, and Y.~Pan, ``A multimedia retrieval
  framework based on semi-supervised ranking and relevance feedback,''
  \emph{IEEE Transactions on Pattern Analysis and Machine Intelligence},
  vol.~34, no.~4, pp. 723--742, 2012.

\bibitem{Zheng2016Personi}
L.~Zheng, Y.~Yang, and A.~G. Hauptmann, ``Person re-identification: Past,
  present and future,'' \emph{arXiv preprint arXiv:1610.02984}, 2016.

\bibitem{li2014deepreid}
W.~Li, R.~Zhao, T.~Xiao, and X.~Wang, ``Deepreid: Deep filter pairing neural
  network for person re-identification,'' in \emph{Proc. CVPR}, 2014, pp.
  152--159.

\bibitem{liao2015person}
S.~Liao, Y.~Hu, X.~Zhu, and S.~Z. Li, ``Person re-identification by local
  maximal occurrence representation and metric learning,'' in \emph{Proc.
  CVPR}, 2015, pp. 2197--2206.

\bibitem{zheng2016mars}
L.~Zheng, Z.~Bie, Y.~Sun, J.~Wang, S.~Wang, C.~Su, and Q.~Tian, ``Mars: A video
  benchmark for large-scale person re-identification,'' in \emph{Proc. ECCV},
  2016, pp. 868--884.

\bibitem{zheng2015scalable}
L.~Zheng, L.~Shen, L.~Tian, S.~Wang, J.~Wang, and Q.~Tian, ``Scalable person
  re-identification: A benchmark,'' in \emph{Proc. ICCV}, 2015, pp. 1116--1124.

\bibitem{zhao2013person}
R.~Zhao, W.~Ouyang, and X.~Wang, ``Person re-identification by salience
  matching,'' in \emph{Proc. ICCV}, 2013, pp. 2528--2535.

\bibitem{zhao2013unsupervised}
------, ``Unsupervised salience learning for person re-identification,'' in
  \emph{Proc. CVPR}, 2013, pp. 3586--3593.

\bibitem{zheng2015query}
L.~Zheng, S.~Wang, L.~Tian, F.~He, Z.~Liu, and Q.~Tian, ``Query-adaptive late
  fusion for image search and person re-identification,'' in \emph{Proc. CVPR},
  2015, pp. 1741--1750.

\bibitem{xia2014supervised}
R.~Xia, Y.~Pan, H.~Lai, C.~Liu, and S.~Yan, ``Supervised hashing for image
  retrieval via image representation learning.'' in \emph{Proc. AAAI}, 2014,
  pp. 2156--2162.

\bibitem{zhao2015deep}
F.~Zhao, Y.~Huang, L.~Wang, and T.~Tan, ``Deep semantic ranking based hashing
  for multi-label image retrieval,'' in \emph{Proc. CVPR}, 2015, pp.
  1556--1564.

\bibitem{lai2015simultaneous}
H.~Lai, Y.~Pan, Y.~Liu, and S.~Yan, ``Simultaneous feature learning and hash
  coding with deep neural networks,'' in \emph{Proc. CVPR}, 2015, pp.
  3270--3278.

\bibitem{lin2015deep}
K.~Lin, H.-F. Yang, J.-H. Hsiao, and C.-S. Chen, ``Deep learning of binary hash
  codes for fast image retrieval,'' in \emph{Proc. CVPR Workshops}, 2015, pp.
  27--35.

\bibitem{zhang2015bit}
R.~Zhang, L.~Lin, R.~Zhang, W.~Zuo, and L.~Zhang, ``Bit-scalable deep hashing
  with regularized similarity learning for image retrieval and person
  re-identification,'' \emph{IEEE Transactions on Image Processing}, vol.~24,
  no.~12, pp. 4766--4779, 2015.

\bibitem{lai2016instance}
H.~Lai, P.~Yan, X.~Shu, Y.~Wei, and S.~Yan, ``Instance-aware hashing for
  multi-label image retrieval,'' \emph{IEEE Transactions on Image Processing},
  vol.~25, no.~6, pp. 2469--2479, 2016.

\bibitem{ahmed2015improved}
E.~Ahmed, M.~Jones, and T.~K. Marks, ``An improved deep learning architecture
  for person re-identification,'' in \emph{Proc. CVPR}, 2015, pp. 3908--3916.

\bibitem{chen2015deep}
S.-Z. Chen, C.-C. Guo, and J.-H. Lai, ``Deep ranking for person
  re-identification via joint representation learning,'' \emph{arXiv preprint
  arXiv:1505.06821}, 2015.

\bibitem{hoffer2015deep}
E.~Hoffer and N.~Ailon, ``Deep metric learning using triplet network,'' in
  \emph{Proc. International Workshop on Similarity-Based Pattern Recognition},
  2015, pp. 84--92.

\bibitem{schroff2015facenet}
F.~Schroff, D.~Kalenichenko, and J.~Philbin, ``Facenet: A unified embedding for
  face recognition and clustering,'' in \emph{Proc. CVPR}, 2015, pp. 815--823.

\bibitem{wang2014learning}
J.~Wang, Y.~Song, T.~Leung, C.~Rosenberg, J.~Wang, J.~Philbin, B.~Chen, and
  Y.~Wu, ``Learning fine-grained image similarity with deep ranking,'' in
  \emph{Proc. CVPR}, 2014, pp. 1386--1393.

\bibitem{sun2014deep}
Y.~Sun, X.~Wang, and X.~Tang, ``Deep learning face representation from
  predicting 10,000 classes,'' in \emph{Proc. CVPR}, 2014, pp. 1891--1898.

\bibitem{sun2014deepid2}
Y.~Sun, Y.~Chen, X.~Wang, and X.~Tang, ``Deep learning face representation by
  joint identification-verification,'' in \emph{Proc. NIPS}, 2014, pp.
  1988--1996.

\bibitem{zheng2016person}
L.~Zheng, H.~Zhang, S.~Sun, M.~Chandraker, Y.~Yang, and Q.~Tian, ``Person
  re-identification in the wild,'' in \emph{Proc. CVPR}, 2017.

\bibitem{zheng2017unlabeled}
Z.~Zheng, L.~Zheng, and Y.~Yang, ``Unlabeled samples generated by gan improve
  the person re-identification baseline in vitro,'' \emph{arXiv preprint
  arXiv:1701.07717}, 2017.

\bibitem{lin2017improving}
Y.~Lin, L.~Zheng, Z.~Zheng, Y.~Wu, and Y.~Yang, ``Improving person
  re-identification by attribute and identity learning,'' \emph{arXiv preprint
  arXiv:1703.07220}, 2017.

\bibitem{sun2017svdnet}
Y.~Sun, L.~Zheng, W.~Deng, and S.~Wang, ``Svdnet for pedestrian retrieval,''
  \emph{arXiv preprint arXiv:1703.05693}, 2017.

\bibitem{su2015multi}
C.~Su, F.~Yang, S.~Zhang, Q.~Tian, L.~S. Davis, and W.~Gao, ``Multi-task
  learning with low rank attribute embedding for person re-identification,'' in
  \emph{Proc. ICCV}, 2015, pp. 3739--3747.

\bibitem{zhao2014learning}
R.~Zhao, W.~Ouyang, and X.~Wang, ``Learning mid-level filters for person
  re-identification,'' in \emph{Proc. CVPR}, 2014, pp. 144--151.

\bibitem{shen2015person}
Y.~Shen, W.~Lin, J.~Yan, M.~Xu, J.~Wu, and J.~Wang, ``Person re-identification
  with correspondence structure learning,'' in \emph{Proc. ICCV}, 2015, pp.
  3200--3208.

\bibitem{prosser2010person}
B.~Prosser, W.-S. Zheng, S.~Gong, T.~Xiang, and Q.~Mary, ``Person
  re-identification by support vector ranking.'' in \emph{Proc. BMVC}, 2010.

\bibitem{ma2012bicov}
B.~Ma, Y.~Su, and F.~Jurie, ``Bicov: a novel image representation for person
  re-identification and face verification,'' in \emph{Proc. BMVC}, 2012.

\bibitem{bazzani2014sdalf}
L.~Bazzani, M.~Cristani, and V.~Murino, ``Sdalf: modeling human appearance with
  symmetry-driven accumulation of local features,'' in \emph{Person
  Re-Identification}.\hskip 1em plus 0.5em minus 0.4em\relax Springer, 2014,
  pp. 43--69.

\bibitem{li2015cross}
S.~Li, M.~Shao, and Y.~Fu, ``Cross-view projective dictionary learning for
  person re-identification,'' in \emph{Proc. AAAI}, 2015, pp. 2155--2161.

\bibitem{ma2012local}
B.~Ma, Y.~Su, and F.~Jurie, ``Local descriptors encoded by fisher vectors for
  person re-identification,'' in \emph{Proc. ECCV}, 2012, pp. 413--422.

\bibitem{koestinger2012large}
M.~Koestinger, M.~Hirzer, P.~Wohlhart, P.~M. Roth, and H.~Bischof, ``Large
  scale metric learning from equivalence constraints,'' in \emph{Proc. CVPR},
  2012, pp. 2288--2295.

\bibitem{hirzer2012person}
M.~Hirzer, P.~M. Roth, and H.~Bischof, ``Person re-identification by efficient
  impostor-based metric learning,'' in \emph{Proc. IEEE International
  Conference on Advanced Video and Signal-Based Surveillance (AVSS)}, 2012, pp.
  203--208.

\bibitem{roth2014mahalanobis}
P.~M. Roth, M.~Hirzer, M.~K{\"o}stinger, C.~Beleznai, and H.~Bischof,
  ``Mahalanobis distance learning for person re-identification,'' in
  \emph{Person Re-Identification}.\hskip 1em plus 0.5em minus 0.4em\relax
  Springer, 2014, pp. 247--267.

\bibitem{yi2014deep}
D.~Yi, Z.~Lei, and S.~Z. Li, ``Deep metric learning for practical person
  re-identification,'' \emph{arXiv preprint arXiv:1407.4979}, 2014.

\bibitem{gray2008viewpoint}
D.~Gray and H.~Tao, ``Viewpoint invariant pedestrian recognition with an
  ensemble of localized features,'' in \emph{Proc. ECCV}, 2008, pp. 262--275.

\bibitem{hirzer2011person}
M.~Hirzer, C.~Beleznai, P.~M. Roth, and H.~Bischof, ``Person re-identification
  by descriptive and discriminative classification,'' in \emph{Scandinavian
  conference on Image analysis}.\hskip 1em plus 0.5em minus 0.4em\relax
  Springer, 2011, pp. 91--102.

\bibitem{ustinova2015multiregion}
E.~Ustinova, Y.~Ganin, and V.~Lempitsky, ``Multiregion bilinear convolutional
  neural networks for person re-identification,'' \emph{arXiv preprint
  arXiv:1512.05300}, 2015.

\bibitem{wang2016joint}
F.~Wang, W.~Zuo, L.~Lin, D.~Zhang, and L.~Zhang, ``Joint learning of
  single-image and cross-image representations for person re-identification,''
  in \emph{Proc. CVPR}, 2016, pp. 1288--1296.

\bibitem{Xiao2016Learning}
T.~Xiao, H.~Li, W.~Ouyang, and X.~Wang, ``Learning deep feature representations
  with domain guided dropout for person re-identification,'' in \emph{Proc.
  CVPR}, 2016, pp. 1249--1258.

\bibitem{Weiss2008Spectral}
Y.~Weiss, A.~Torralba, and R.~Fergus, ``Spectral hashing.'' in \emph{Proc.
  NIPS}, 2008, pp. 1753--1760.

\bibitem{Gong2011Iterative}
Y.~Gong and S.~Lazebnik, ``Iterative quantization: a procrustean approach to
  learning binary codes for large-scale image retrieval.'' in \emph{Proc.
  CVPR}, 2011, pp. 2916--2929.

\bibitem{Wang2010Semi}
J.~Wang, S.~Kumar, and S.~F. Chang, ``Semi-supervised hashing for scalable
  image retrieval,'' in \emph{Proc. CVPR}, 2010, pp. 3424--3431.

\bibitem{Norouzi2011Minimal}
M.~Norouzi, ``Minimal loss hashing for compact binary codes.'' in \emph{Proc.
  ICML}, 2011, pp. 353--360.

\bibitem{yang2015robust}
Y.~Yang, F.~Shen, H.~T. Shen, H.~Li, and X.~Li, ``Robust discrete spectral
  hashing for large-scale image semantic indexing,'' \emph{IEEE Transactions on
  Big Data}, vol.~1, no.~4, pp. 162--171, 2015.

\bibitem{Yang2016Zero}
Y.~Yang, W.~Chen, Y.~Luo, F.~Shen, J.~Shao, and H.~T. Shen, ``Zero-shot hashing
  via transferring supervised knowledge,'' in \emph{Proc. ACM International
  Conference on Multimedia}, 2016, pp. 1286--1295.

\bibitem{Liu2012Supervised}
W.~Liu, J.~Wang, R.~Ji, Y.~G. Jiang, and S.~F. Chang, ``Supervised hashing with
  kernels,'' in \emph{Proc. CVPR}, 2012, pp. 2074--2081.

\bibitem{krizhevsky2012imagenet}
A.~Krizhevsky, I.~Sutskever, and G.~E. Hinton, ``Imagenet classification with
  deep convolutional neural networks,'' in \emph{Proc. NIPS}, 2012, pp.
  1097--1105.

\bibitem{Chang2016Convex}
X.~Chang, F.~Nie, Y.~Yang, C.~Zhang, and H.~Huang, ``Convex sparse pca for
  unsupervised feature learning,'' \emph{ACM Transactions on Knowledge
  Discovery from Data}, vol.~11, no.~1, pp. 3:1--3:16, 2016.

\bibitem{chang2016semisupervised}
X.~Chang and Y.~Yang, ``Semisupervised feature analysis by mining correlations
  among multiple tasks,'' \emph{IEEE Transactions on Neural Networks and
  Learning Systems}, 2016.

\bibitem{arandjelovic2012multiple}
R.~Arandjelovic and A.~Zisserman, ``Multiple queries for large scale specific
  object retrieval.'' in \emph{Proc. BMVC}, 2012, pp. 1--11.

\bibitem{farenzena2010person}
M.~Farenzena, L.~Bazzani, A.~Perina, V.~Murino, and M.~Cristani, ``Person
  re-identification by symmetry-driven accumulation of local features,'' in
  \emph{Proc. CVPR}, 2010, pp. 2360--2367.

\bibitem{Wang2010Sequential}
J.~Wang, S.~Kumar, and S.~F. Chang, ``Sequential projection learning for
  hashing with compact codes,'' in \emph{Proc. ICML}, 2010, pp. 1127--1134.

\bibitem{Heo2012Spherical}
J.~P. Heo, Y.~Lee, J.~He, S.~F. Chang, and S.~E. Yoon, ``Spherical hashing,''
  in \emph{Proc. CVPR}, 2012, pp. 2957--2964.

\bibitem{Jin2014Density}
Z.~Jin, C.~Li, Y.~Lin, and D.~Cai, ``Density sensitive hashing.'' \emph{IEEE
  Transactions on Cybernetics}, vol.~44, no.~8, pp. 1362--1371, 2014.

\bibitem{shen2015supervised}
F.~Shen, C.~Shen, W.~Liu, and H.~Tao~Shen, ``Supervised discrete hashing,'' in
  \emph{Proc. CVPR}, 2015, pp. 37--45.

\bibitem{Jia2014Caffe}
Y.~Jia, E.~Shelhamer, J.~Donahue, S.~Karayev, J.~Long, R.~Girshick,
  S.~Guadarrama, and T.~Darrell, ``Caffe: Convolutional architecture for fast
  feature embedding,'' in \emph{Proc. ACM International Conference on
  Multimedia}, 2014, pp. 675--678.

\bibitem{deng2009imagenet}
J.~Deng, W.~Dong, R.~Socher, L.-J. Li, K.~Li, and L.~Fei-Fei, ``Imagenet: A
  large-scale hierarchical image database,'' in \emph{Proc. CVPR}, 2009, pp.
  248--255.

\bibitem{weinberger2005distance}
K.~Q. Weinberger, J.~Blitzer, and L.~K. Saul, ``Distance metric learning for
  large margin nearest neighbor classification,'' in \emph{Proc. NIPS}, 2005,
  pp. 1473--1480.

\bibitem{davis2007information}
J.~V. Davis, B.~Kulis, P.~Jain, S.~Sra, and I.~S. Dhillon,
  ``Information-theoretic metric learning,'' in \emph{Proc. ICML}, 2007, pp.
  209--216.

\bibitem{wu2016personnet}
L.~Wu, C.~Shen, and A.~v.~d. Hengel, ``Personnet: Person re-identification with
  deep convolutional neural networks,'' \emph{arXiv preprint arXiv:1601.07255},
  2016.

\bibitem{Su2016Deep}
C.~Su, S.~Zhang, J.~Xing, W.~Gao, and Q.~Tian, ``Deep attributes driven
  multi-camera person re-identification,'' in \emph{Proc. ECCV}, 2016, pp.
  475--491.

\bibitem{martinel2016temporal}
N.~Martinel, A.~Das, C.~Micheloni, and A.~K. Roy-Chowdhury, ``Temporal model
  adaptation for person re-identification,'' \emph{arXiv preprint
  arXiv:1607.07216}, 2016.

\bibitem{liu2016end}
H.~Liu, J.~Feng, M.~Qi, J.~Jiang, and S.~Yan, ``End-to-end comparative
  attention networks for person re-identification,'' \emph{arXiv preprint
  arXiv:1606.04404}, 2016.

\end{thebibliography}
\begin{IEEEbiography}[{\includegraphics[width=1in,height=1.25in,clip,keepaspectratio]{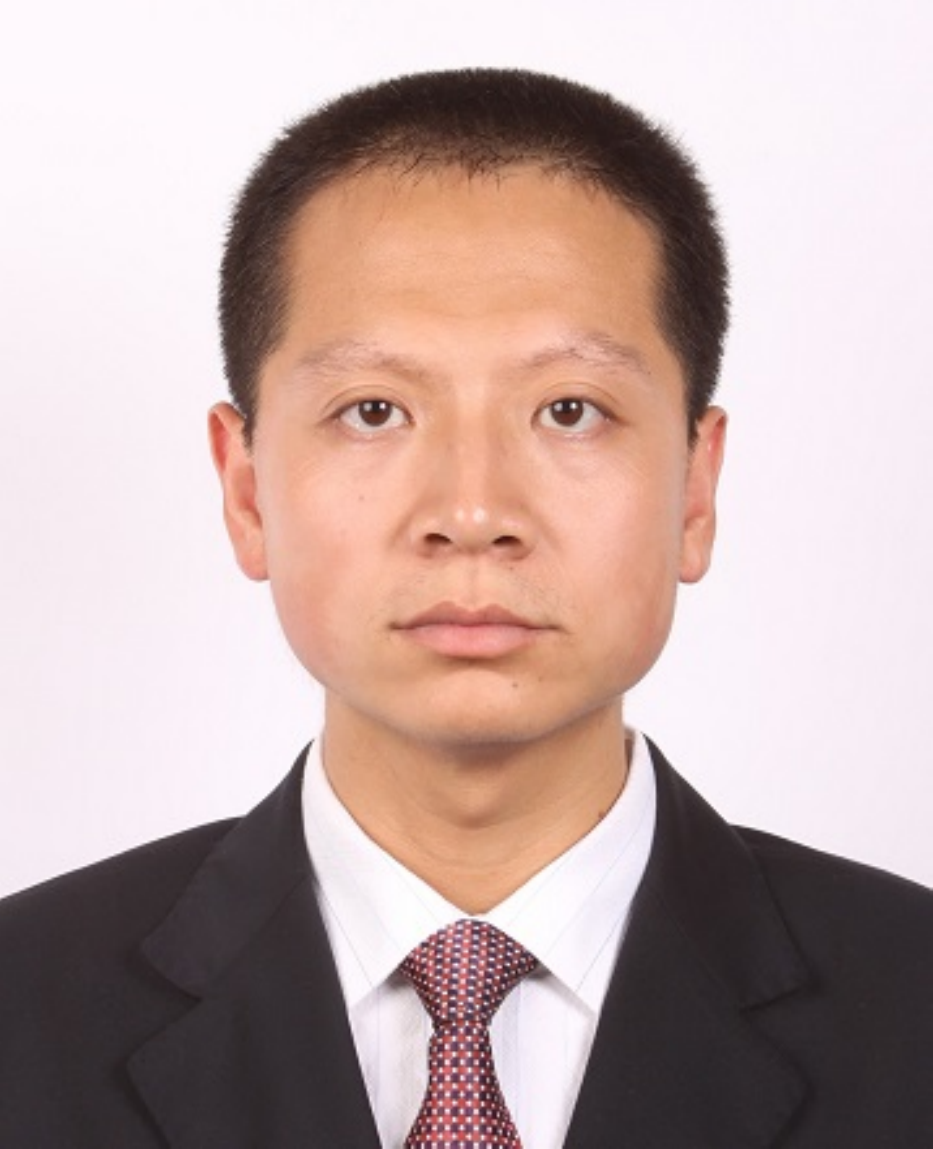}}]{Fuqing Zhu}
received his B.E. and M.S. degree from Dalian Jiaotong University, China, in 2010 and 2013, respectively. Currently, he is seeking his Ph.D. degree in School of Information and Communication Engineering at Dalian University of Technology, China. His research interests include multimedia retrieval, image classification and person re-identification.
\end{IEEEbiography}
\begin{IEEEbiography}[{\includegraphics[width=1in,height=1.25in,clip,keepaspectratio]{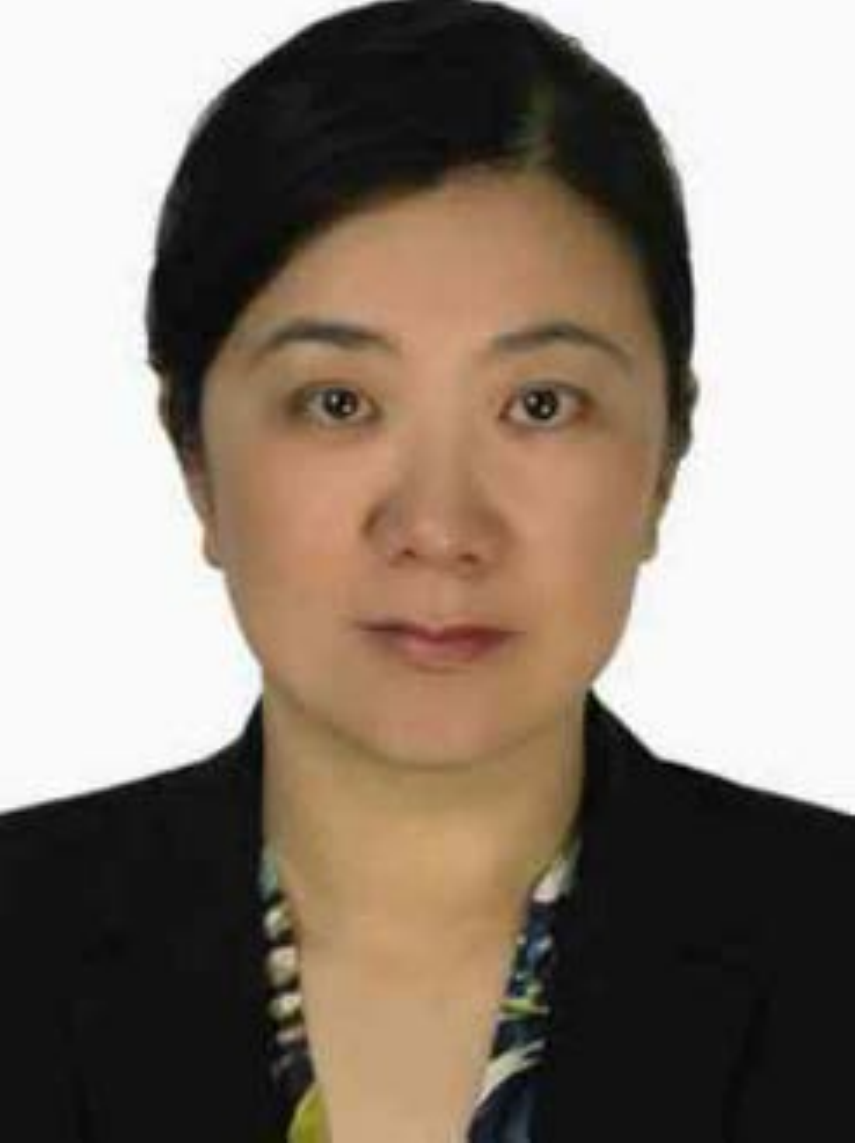}}]{Xiangwei Kong}
received her Ph.D. degree in Management Science and Engineering from Dalian University of Technology, China, in 2003. From 2006 to 2007, she  was a visiting scholar in Department of Computer Science at Purdue University, USA. From 2014 to 2015, she was a senior research scientist in Department of Computer Science at New York University, USA. She is currently a professor in School of Information and Communication Engineering, and the director of research center of multimedia information processing and security at Dalian University of Technology, China. She has published 4 edited books and more than 185 research papers in refereed international journals and conferences in the areas of cross-modal retrieval, multimedia information security, knowledge mining and business intelligence.
\end{IEEEbiography}
\begin{IEEEbiography}[{\includegraphics[width=1in,height=1.25in,clip,keepaspectratio]{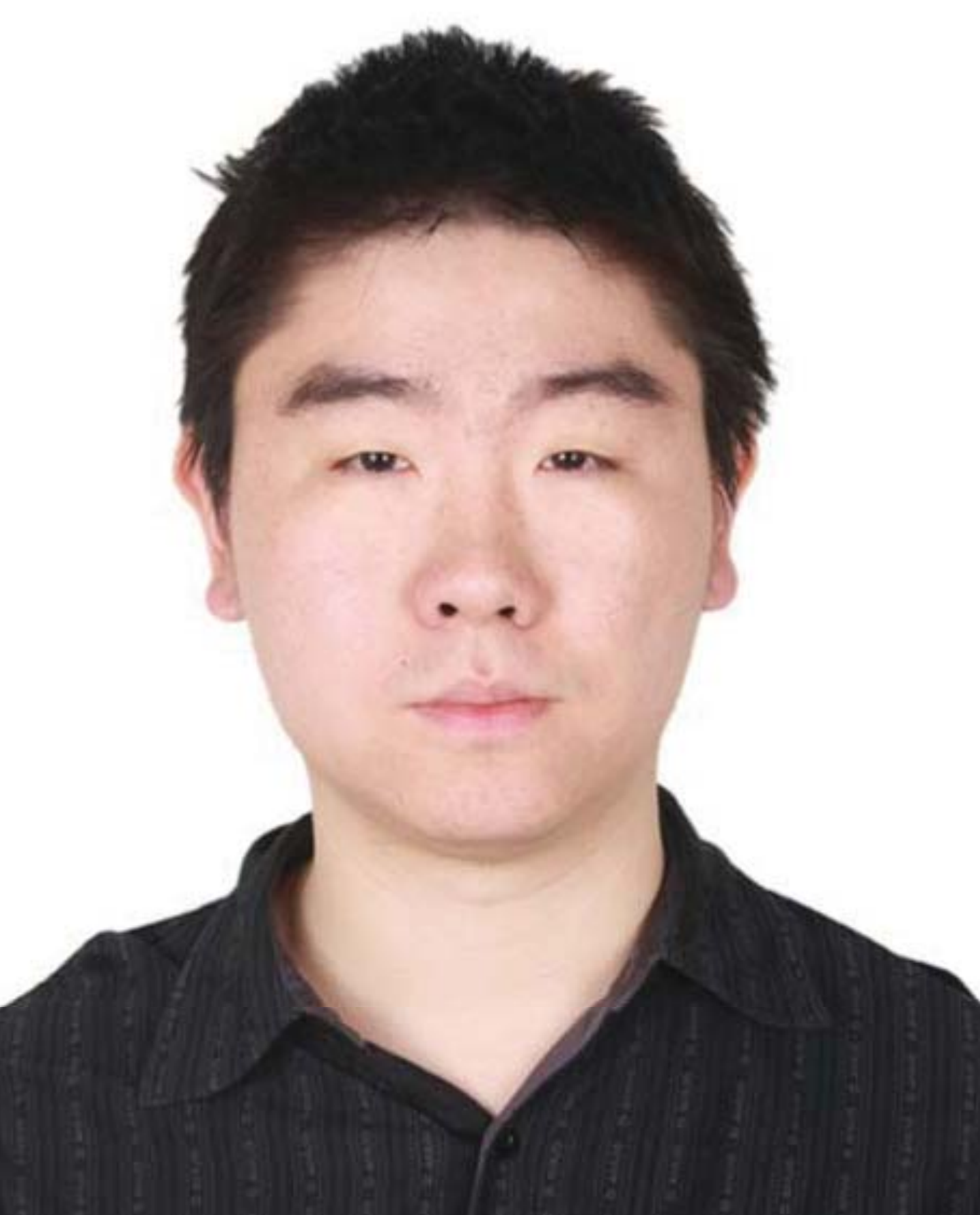}}]{Liang Zheng}
received the Ph.D. degree in Electronic Engineering from Tsinghua University, China, in 2015, and the  B.E. degree in Life Science from Tsinghua University, China, in 2010. He was a postdoc researcher in University of Texas at San Antonio, USA. He is currently a postdoc researcher in Quantum Computation and Intelligent Systems, University of Technology Sydney, Australia. His research interests include image retrieval, classification, and person re-identification.
\end{IEEEbiography}
\begin{IEEEbiography}[{\includegraphics[width=1in,height=1.25in,clip,keepaspectratio]{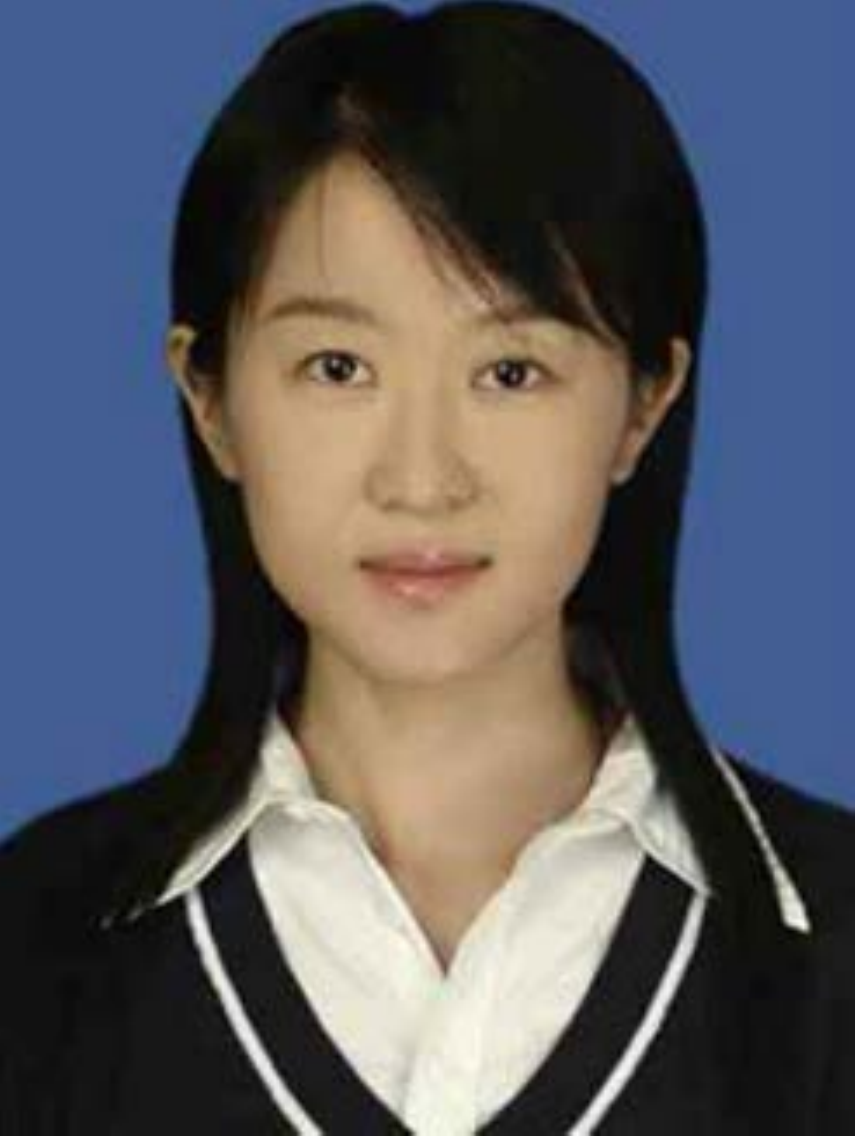}}]{Haiyan Fu}
received her Ph.D. degree from Dalian University of Technology, China, in 2014. She is currently an associate professor in School of Information and Communication Engineering at Dalian University of Technology, China. Her research interests are in the areas of image retrieval and computer vision.
\end{IEEEbiography}
\vfill
\begin{IEEEbiography}[{\includegraphics[width=1in,height=1.25in,clip,keepaspectratio]{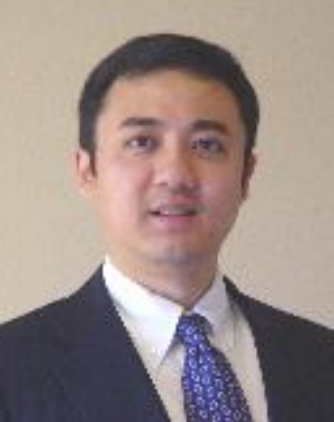}}]{Qi Tian}
(M'96-SM'03-F'16) received his Ph.D. in ECE from University of Illinois at Urbana-Champaign (UIUC) in 2002 and received his B.E. in Electronic Engineering from Tsinghua University in 1992 and M.S. in ECE from Drexel University in 1996, respectively. He is currently a Full Professor in the Department of Computer Science, the University of Texas at San Antonio (UTSA). He was a tenured Associate Professor from 2008-2012 and a tenure-track Assistant Professor from 2002-2008. During 2008-2009, he took one-year Faculty Leave at Microsoft Research Asia (MSRA) as Lead Researcher in the Media Computing Group. Dr. Tian’s research interests include multimedia information retrieval, computer vision, pattern recognition and bioinformatics and published over 350 refereed journal and conference papers. He was the co-author of a \textbf{Best Paper} in ACM ICMR 2015, a \textbf{Best Paper} in PCM 2013, a \textbf{Best Paper} in MMM 2013, a \textbf{Best Paper} in ACM ICIMCS 2012, a \textbf{Top 10\% Paper Award} in MMSP 2011, a \textbf{Best Student Paper} in ICASSP 2006, and co-author of a \textbf{Best Student Paper Candidate} in ICME 2015, and a \textbf{Best Paper Candidate} in PCM 2007. Dr. Tian research projects are funded by ARO, NSF, DHS, Google, FXPAL, NEC, SALSI, CIAS, Akiira Media Systems, HP, Blippar and UTSA. He received 2014 Research Achievement Awards from College of Science, UTSA. He received \textbf{2016 UTSA Innovation Award} and \textbf{2010 ACM Service Award}. He is the associate editor of IEEE Transactions on Multimedia (TMM), IEEE Transactions on Circuits and Systems for Video Technology (TCSVT), ACM Transactions on Multimedia Computing, Communications, and Applications (TOMM), Multimedia System Journal (MMSJ), and in the Editorial Board of Journal of Multimedia (JMM) and Journal of Machine Vision and Applications (MVA).  Dr. Tian is the Guest Editor of IEEE Transactions on Multimedia, Journal of Computer Vision and Image Understanding, etc.
\end{IEEEbiography}
\vfill

\end{document}